\documentclass[10pt,twocolumn,letterpaper]{article}

 \usepackage[pagenumbers]{iccv} 

\definecolor{iccvblue}{rgb}{0.21,0.49,0.74}

\definecolor{mygray}{gray}{.9}
\newcommand{\pub}[1]{{\tiny{[{#1}]}}}

\usepackage[pagebackref,breaklinks,colorlinks,allcolors=iccvblue]{hyperref}
\usepackage{multirow}
\usepackage{colortbl}
\usepackage{bm}
\usepackage{algorithm}
\usepackage{algorithmicx}
\usepackage{algpseudocode}
\usepackage{xcolor}
\usepackage{textcomp}
\usepackage{booktabs}
\usepackage{multicol}
\usepackage{makecell}

\def\confName{ICCV}
\def\confYear{2025}

\title{FedPCA: Noise-Robust Fair Federated Learning via Performance-Capacity Analysis}

\author{
	Nannan Wu$^{1,2}$\thanks{Preprint version. Additional information and modifications will be included in future versions.}\quad
	Zengqiang Yan$^{2}$\quad
	Nong Sang$^{2}$\quad
	Li Yu$^{2}$\quad
	Chang Wen Chen$^{1}$
	\\ $^{1}$ The Hong Kong Polytechnic University\quad$^{2}$ Huazhong University of Science and Technology\\
}

\begin{document}
	\maketitle	
	\begin{abstract}
		
		Training a model that effectively handles both common and rare data—\textit{i.e.}, achieving performance fairness—is crucial in federated learning (FL). While existing fair FL methods have shown effectiveness, they remain vulnerable to mislabeled data. Ensuring robustness in fair FL is therefore essential. However, fairness and robustness inherently compete, which causes robust strategies to hinder fairness. In this paper, we attribute this competition to the homogeneity in loss patterns exhibited by rare and mislabeled data clients, preventing existing loss-based fair and robust FL methods from effectively distinguishing and handling these two distinct client types. To address this, we propose performance-capacity analysis, which jointly considers model performance on each client and its capacity to handle the dataset, measured by loss and a newly introduced feature dispersion score. This allows mislabeled clients to be identified by their significantly deviated performance relative to capacity while preserving rare data clients. Building on this, we introduce FedPCA, an FL method that robustly achieves fairness. FedPCA first identifies mislabeled clients via a Gaussian Mixture Model on loss-dispersion pairs, then applies fairness and robustness strategies in global aggregation and local training by adjusting client weights and selectively using reliable data. Extensive experiments on three datasets demonstrate FedPCA's effectiveness in tackling this complex challenge. Code will be publicly available upon acceptance.
		
	\end{abstract}

	\section{Introduction}
	\label{sec:intro}
	
	\begin{figure}[!t] 
		\centering
		\includegraphics[width=1.0\columnwidth]{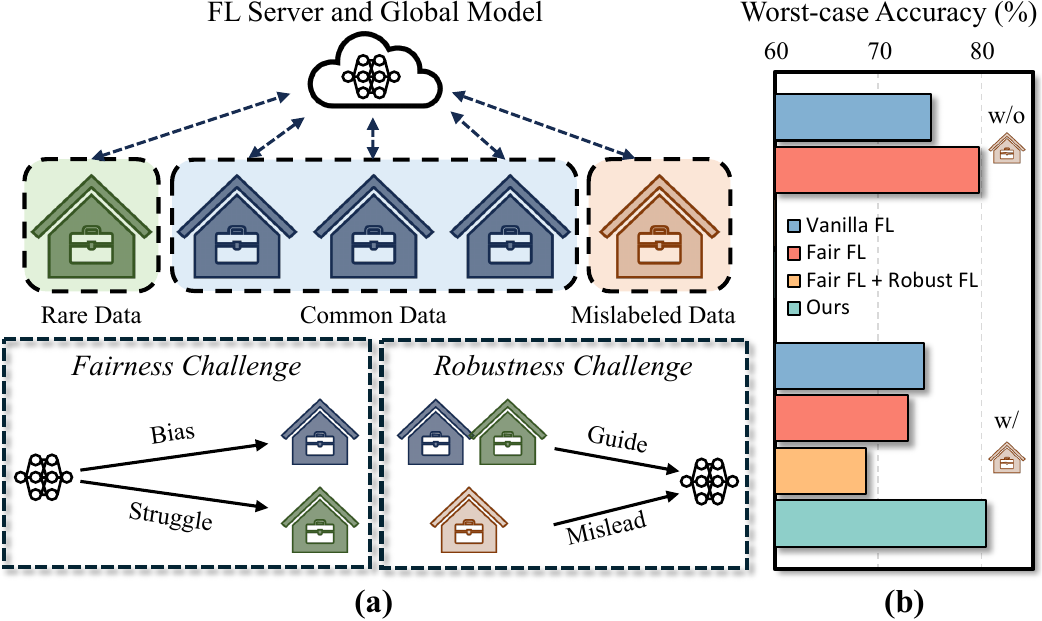}
		\caption{\textbf{(a)} Our setting involves both fairness and robustness challenges. The former arises from imbalanced data scales across diverse data distributions, while the latter stems from clients with mislabeled data. \textbf{(b)} In the absence of mislabeled data, fairness can be improved using existing fair FL methods like \cite{FedISM}. However, these methods remain highly vulnerable to mislabeled data, even when combined with state-of-the-art robust FL techniques \cite{FedNoRo}. Our proposed method effectively addresses this hybrid challenge, achieving fairness in a robust manner. Results are obtained from experiments conducted on RSNA ICH.}
		\label{fig:background}
	\end{figure}

	With growing privacy concerns, Federated Learning (FL) \cite{FedAvg} has emerged as a promising paradigm for model training, particularly in privacy-sensitive domains like healthcare \cite{dou2021federated}, finance \cite{imteaj2022leveraging}, and IoT \cite{mills2019communication}. By incorporating participants with heterogeneous data distributions, FL expands data scale and diversity, potentially improving generalization. However, a key challenge is ensuring the model can effectively handle such diverse client data—namely, achieving Rawlsian Max-Min fairness \cite{rawls2001justice} in client performance.
	This \textit{fairness challenge} arises because real-world data scales across distributions are often highly imbalanced \cite{FedISM,he2009learning}. Consequently, models trained on such biased data tend to exhibit performance disparities \cite{q-FedAvg,FedISM}, excelling on distributions with abundant data (\textit{i.e.,} common data) while struggling with those containing fewer samples (\textit{i.e.,} rare data), as illustrated in Fig. \ref{fig:background} (a). This limits FL's applicability, as deploying such models in environments dominated by insufficiently learned rare data is ineffective \cite{q-FedAvg}.
	To address this, existing fair FL methods prioritize learning from clients with lower performance \cite{q-FedAvg,FedCE,FedGA,FedISM}, demonstrating effectiveness, as shown in Fig. \ref{fig:background} (b).
	
	Despite enhancing fairness, these methods rely on the unrealistic assumption that all clients provide correctly labeled data for performance evaluation. However, in FL's decentralized setting, ensuring label purity across all participants is inherently difficult \cite{huang2024federated}. Label quality is often compromised by annotation errors \cite{han2018co,liu2020early,liu2022adaptive} or adversarial manipulations \cite{wang2020attack}, introducing an additional \textit{robustness challenge}, as illustrated in Fig. \ref{fig:background} (a). Our empirical results in Fig. \ref{fig:background} (b) show that existing fair FL methods are highly vulnerable to mislabeled data, suffering severe performance degradation, even falling below the vanilla FL baseline \cite{FedAvg}. A straightforward idea is to integrate these methods with well-established robust FL techniques \cite{Median,RFA,FedCorr,FedNoRo}. However, as shown in Fig. \ref{fig:background} (b), this combination remains ineffective in tackling the hybrid challenge. In short, robustly achieving fairness under label noise remains an open problem, which we investigate in this paper.
	
	The core difficulty of this hybrid problem lies in the inherent competition between fairness and robustness for the global model \cite{li2021ditto,zhou2024h}.
	In this paper, we investigate the root cause of this competition as the similar loss patterns exhibited by rare and mislabeled clients relative to common data clients. The former occurs because rare data is insufficiently learned, while the latter arises due to the early-learning phenomenon \cite{liu2020early,liu2022adaptive}. These properties are leveraged in fair FL \cite{mohri2019agnostic,FedISM,q-FedAvg} and robust FL \cite{FedCorr,FedNoRo,li2024feddiv} to address each problem separately. However, in this hybrid setting, existing solutions that rely heavily on loss metrics fail to distinguish between rare and mislabeled clients, assigning them similar importance. As illustrated in Fig. \ref{fig:weights_pre}, fair FL, prioritizing rare data clients, inadvertently amplify the influence of mislabeled clients, compromising robustness. While incorporating robust FL techniques can down-weight mislabeled clients, it risks neglecting rare data clients, undermining fairness. This underscores the need for a novel approach that harmonizes fairness and robustness, rather than treating them as competing trade-offs.
	
	Mitigating such competition requires precisely filtering out mislabeled clients while preserving those with rare data, leading to a fundamental question:
	1) \textbf{\textit{What distinguishes mislabeled data clients?}}
	Inspired by real-world learning, we argue that the defining characteristic is not an absolutely high loss—both mislabeled and rare-data clients exhibit this—but rather a relatively high loss compared to the model's capacity to handle that dataset. To capture this distinction, we propose performance-capacity analysis, where label errors are identified when model performance significantly deviates from its expected capacity.
	This raises another question:
	2)	\textbf{\textit{How can we precisely estimate the model's capacity for a given dataset?}}
	To address this, we introduce a label-free metric: the feature dispersion score, inspired by recent work on out-of-distribution error prediction \cite{xie2024importance}. This metric quantifies how well middle-layer features extracted by models separate across classes. Empirically and theoretically, it exhibits an inverse correlation with the actual prediction error \cite{xie2024importance}, providing a complementary perspective to the loss metric. 
	By combining the dispersion score and loss metric, we establish a more reliable approach to distinguishing clients and mitigating competition.
	
	With these two complementary metrics, the next question arises:
	3)	\textbf{\textit{How can we integrate these two metrics into FL to achieve noise-robust fairness?}} To this end, we propose \textbf{FedPCA} (Federated Learning through Performance-Capacity Analysis). The key idea is to leverage these two metrics to filter out mislabeled data and guide fair learning. We achieve this by first grouping clients using a Gaussian Mixture Model based on their two complementary metrics. Clients with anomalous loss-feature dispersion pairs are identified as likely containing mislabeled data. For these clients, we introduce two data selection strategies:
	1) completely discarding data from identified mislabeled clients, or
	2) selectively retaining high-confidence samples, using confidently predicted labels for training. This improves local training reliability. At the global aggregation stage, we incorporate both fairness and robustness by assigning greater weight to clients with lower feature dispersion scores and a higher proportion of reliable labels, ensuring that data from highly reliable clients with greater learning difficulty is appropriately prioritized. Our solution is simple yet effective, requiring only minor modifications to local training data selection and aggregation weights compared to FedAvg \cite{FedAvg}. Its effectiveness is validated through extensive experiments on three datasets.
	
	In summary, our contributions are four folds:
	\begin{itemize}
		\item We tackle the hybrid challenge of robustly achieving fairness, providing insights into the competition between fairness and robustness—namely, that clients with distinct characteristics exhibit similar loss patterns.
		
		\item To resolve this, we propose performance-capacity analysis framework, designing two complementary metrics that enable the differentiation of mislabeled clients through anomalous metric pairs while preserving other clients.
		
		\item Building on these metrics, we propose \textbf{FedPCA}, a simple yet effective FL algorithm that integrates fairness and robustness in both local training and global aggregation.
		
		\item We validate our approach on representative image classification tasks, demonstrating its effectiveness in address this problem.
	\end{itemize}

	\section{Related work}  \label{sec:Related work}
	
	\subsection{Performance-Fair Federated Learning}
	Performance fairness, which aims to ensure a high lower-bound performance \cite{mohri2019agnostic,FedISM} and minimize performance disparity \cite{q-FedAvg} across distributions, is crucial in FL as it determines the model's applicability.
	To address this, current research typically establishes a fairness surrogate and adjusts the training focus dynamically based on its discrepancy with the current state. For instance, q-FedAvg \cite{q-FedAvg} seeks to equalize training loss across clients by assigning higher weights to those with greater loss values. FedGA \cite{FedGA}, FedCE \cite{FedCE}, and FedISM \cite{FedISM} further extend this strategy by exploring other surrogates based on metrics such as generalization gap, task accuracy, and loss surface sharpness. Despite their effectiveness, these methods lack robustness when encountering clients with label noise \cite{li2021ditto}, as the adopted surrogate fails to exclude such clients effectively, leading to misplaced training focus.
	
	\subsection{Noise-Robust Federated Learning}
	As a decentralized training paradigm, FL cannot ensure the reliability of all participants' data, leading to the possibility of clients containing incorrect labels \cite{huang2024federated}. Previous research shows that such erroneous labels can degrade performance \cite{han2018co}, underscoring the need for noise-robust FL. Current solutions to this problem generally fall into two categories: (1) general robust FL against Byzantine attacks \cite{Median,RFA}, and (2) specialized robust FL targeting label noise \cite{FedCorr,FedNoRo,li2024feddiv}. However, existing methods heavily rely on the \textit{i.i.d.} assumption on data or related performance statistics. When data scales are imbalanced across distributions, these assumptions fail, compromising effectiveness. For instance, robust methods may erroneously exclude rare data clients simply because they share certain metric characteristics used to identify mislabeled data clients \cite{wang2020attack}.
	
	\subsection{Fair and Robust Federated Learning}
	Both fairness and robustness are crucial to FL's trustworthiness, yet research that jointly addresses both issues is limited. Xu et al. \cite{xu2020reputation} consider both aspects, but their focus is on collaborative fairness, which is inherently linked to robustness. Ditto \cite{li2021ditto} marks an initial attempt to address robustness alongside performance fairness by training a personalized model instead of a global one. H-nobs \cite{zhou2024h} actively combines robust aggregation with q-FedAvg \cite{q-FedAvg} to tackle this problem.
	However, the competition between these two goals forces existing methods to rely on trade-offs rather than fully achieving both.

	\section{Methodology}
	\subsection{Preliminaries} \label{sec:preliminaries}
	We consider a cross-silo FL environment with $K$ clients participating in a classification task, with input space $\mathcal{X}$ and label space $\mathcal{Y} = [C]$\footnote{$[C]$ represents the set $\{1,2,\cdots,C\}$}. Each client $k$ has a dataset $D_k=\{(\boldsymbol x_i \in \mathcal{X}, y_i \in \mathcal{Y})\}_{i=1}^{N_k}$ of size $N_k$. For simplicity and without loss of generality, we assume that each dataset $D_k$ is drawn from either a common data distribution $\mathcal{P}_1$ or a rare data distribution $\mathcal{P}_2$, with most clients sampling from $\mathcal{P}_1$. We also consider the presence of label noise, meaning that these ideal datasets are not directly accessible. Instead, each client observes a noisy dataset $\bar{D}_k=\{(\boldsymbol x_i, \bar{y}_i)\}_{i=1}^{N_k}$. 	For some clients, we have that $\bar{y} \not\equiv y$.
	
	The objective is to train a global model with parameters $\boldsymbol{\theta}$, consisting of a feature extractor $f: \mathcal{X} \rightarrow \mathcal{Z}$ and a classifier $g: \mathcal{Z} \rightarrow \Delta^{C-1}$, where $\mathcal{Z}$ denotes the feature space and $\Delta$ represents the probability simplex, such that:
	\begin{equation} \label{eq:objective}
		\min_{\boldsymbol{\theta}} \max_{m \in \{1,2\}} \mathbb{E}_{(\boldsymbol{x}, y) \sim \mathcal{P}_m} [\ell(g \circ f(\boldsymbol{x}; \boldsymbol{\theta}), y)],
	\end{equation}
	where $\ell: \Delta^{C-1} \times \mathcal{Y} \rightarrow \{0\} \cup \mathbb{R}_+$ denotes a loss function (\textit{e.g.}, cross-entropy loss) measuring the discrepancy between predictions and true labels. This objective aims to robustly achieve Rawlsian Max-Min fairness \cite{rawls2001justice} in the presence of label noise. Tab. \ref{tab:setting} summarizes a comparison with related settings, highlighting our more relaxed assumptions.
	
	\begin{table}[!t]
		\centering
		\caption{Comparison of settings across different FL studies.}
		\label{tab:setting}
		\resizebox{0.9\columnwidth}{!}{
			\begin{tabular}{lccc}
				\toprule
				\textbf{Setting} & \makecell[c]{Imbalanced \\ Distributions} & \makecell[c]{Incorrect \\ Labels} & Objective \\ 
				\midrule
				Fair FL         & \textcolor{green}{\checkmark} & \textcolor{red}{\texttimes} & Achieve fairness \\
				Robust FL       & \textcolor{red}{\texttimes} & \textcolor{green}{\checkmark} & Achieve robustness \\
				\midrule
				\textbf{This paper} & \textcolor{green}{\checkmark} & \textcolor{green}{\checkmark} & Robustly achieve fairness \\
				\bottomrule
			\end{tabular}
		}
	\end{table}

	\subsection{Problem Analysis} \label{sec:PA}
\begin{figure}[!t] 
		\centering
		\includegraphics[width=1.0\columnwidth]{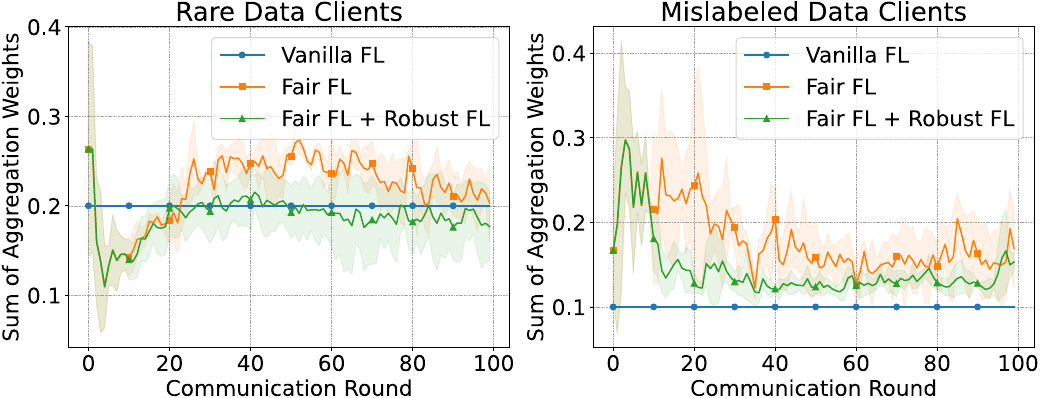}
		\caption{Comparison of aggregation weights assigned to rare and mislabeled data clients for vanilla FL \cite{FedAvg}, fair FL \cite{FedISM}, and its combination with robust FL \cite{FedNoRo}. The transparent area represents the standard deviation. Experiments are conducted on RSNA ICH.}
		\label{fig:weights_pre}
	\end{figure}

	\noindent\textbf{Competition between Fairness and Robustness.}
	In the presence of label noise, Eq. \ref{eq:objective} defines a dual objective: ensuring robustness and achieving fairness. Robustness requires down-weighting clients with incorrect labels, while fairness demands greater attention to rare data clients. However, it can be observed from our empirical study that these objectives inherently compete in existing solution frameworks. As shown in Fig. \ref{fig:weights_pre}, this competition is reflected in the aggregation weights assigned to rare and mislabeled clients. Compared to the vanilla FL baseline \cite{FedAvg}, fair FL \cite{FedISM} prioritizes rare data clients but inadvertently assigns excessive attention to mislabeled ones, improving fairness at the expense of robustness. Integrating robust techniques \cite{FedNoRo} into fair FL mitigates this issue but simultaneously reduces attention to rare data clients, improving robustness at the cost of fairness. In summary, fairness and robustness inherently compete, making it challenging to improve both simultaneously. To address this problem, a harmonized approach that robustly achieves fairness is needed.
	
	\noindent\textbf{Homogeneity in Loss Leads to Competition.}
	To resolve this competition, it is essential to first understand its root cause. In this paper, we find that the competition between fairness and robustness arises because rare and mislabeled data clients exhibit similar loss trends relative to common clients. Loss or it variants, which quantify the discrepancy between model predictions and given labels, are widely used in FL research focusing separately on fairness \cite{q-FedAvg,FedGA,FedCE,FedISM} or robustness \cite{FedCorr,FedNoRo}. In fair FL, rare data clients typically show higher loss than common ones, justifying higher weight assignments to mitigate performance disparities. Conversely, robust FL leverages the early-learning phenomenon \cite{han2018co,liu2022adaptive,liu2020early}, where mislabeled data leads to significantly higher loss, enabling the filtering of high-loss clients to enhance robustness.
	
	However, when fairness and robustness must be addressed simultaneously, this metric faces a critical limitation: both rare and mislabeled data clients exhibit higher loss than common data clients. Moreover, as imbalance severity and noise rate vary, their relative ranking is not fixed, and their loss distributions can even theoretically overlap entirely, making them indistinguishable. This loss homogeneity forces them to be assigned similar importance, creating a competitive dynamic. As shown in Fig. \ref{fig:weights_pre}, up-weighting high-loss clients risks amplifying erroneous data, while down-weighting them overlooks valuable rare data.

	\subsection{Performance-Capacity Analysis}
	\noindent\textbf{Mitigating Competition by Incorporating Capacity.} 
	To address this competition, a complementary perspective is needed—one that accurately differentiates mislabeled clients while preserving rare data. A key question thus arises: What distinguishes mislabeled clients? We draw inspiration from real-world learning environments. If a student what to question a teacher's answer, they should not rely solely on a low exam score but also consider their own capacity to solve similar problems. A teacher's answer is likely incorrect only if the student's performance significantly deviates from their expected capacity. Similarly, identifying mislabeled clients requires considering the model's capacity to handle data and integrating it with model performance (\textit{i.e.}, loss) for reliable client identification. This approach effectively distinguishes rare data clients from mislabeled ones.
	
	The next question is: How can we estimate model capacity? This should be done without relying on labels, ensuring it complements performance in assessing label correctness. Recent research on out-of-distribution error prediction suggests that feature dispersion correlates strongly with true prediction error \cite{xie2024importance}. This relationship is theoretically supported by the Bayes error rate, which establishes a lower bound for classification error based on feature distances. In binary classification:
	\begin{equation} \label{eq:bayes_binary}
		E_{\text {bayes }} \leq \frac{2 P(y=1) P(y=2)}{1 + P(y=1) P(y=2) \delta},
	\end{equation}
	where $P(y=i)$ denotes the class prior, and $\delta$ represents the feature distance between classes \cite{devijver1982pattern,tumer2003bayes}. For multi-class settings with $C > 2$, the upper bound generalizes to  \cite{garber1988bounds}:
	\begin{equation} \label{eq:bayes_multi}
		\begin{aligned}
			E_{\text {bayes }}^C \leq \min _{\alpha \in\{0,1\}} \Bigg( & \frac{1}{C-2 \alpha} \sum_{i=1}^C \left(1 - P\left(y=i\right)\right) E_{\text {bayes}; i}^{C-1} \\
			& + \frac{1 - \alpha}{C-2 \alpha} \Bigg).
		\end{aligned}
	\end{equation}
	These findings suggest that increased feature separation is possibly associated with lower prediction error, a conclusion also empirically supported in representation learning studies aiming for class-separated features \cite{chen2020simple,khosla2020supervised,oord2018representation}. To quantify this separation, we employ the dispersion score \cite{xie2024importance}, which computes the average distance between cluster centroids in the feature space $\mathcal{Z}$ for the dataset of each client $k$, as:
	\begin{equation} \label{eq:score}
		S(\bar{D}_k) = \log \frac{\sum_{j=1}^{C} n_{k,j} \cdot \|\bar{\boldsymbol{\mu}}_k - \tilde{\boldsymbol{\mu}}_{k,j}\|_2^2}{C-1},
	\end{equation}
	where $n_{k,j}$ represents the sample count for cluster $j$ in client $k$, $\tilde{\boldsymbol{\mu}}_{k,j}$ is the feature centroid for that cluster, and $\bar{\boldsymbol{\mu}}_k$ denotes the mean of all features $f(\boldsymbol{x})$. This metric, capturing model capacity independently of labels, offers an orthogonal perspective to the loss metric.
	
	\noindent\textbf{Integrating Performance and Capacity.} \label{sec:IPC}
	We jointly analyze performance and capacity to differentiate mislabeled clients, which is fundamental for robustly achieving fairness. Performance is measured using the loss metric (\textit{e.g.,} cross-entropy loss), while capacity is measured via the feature dispersion score. These two perspectives are complementary: loss captures the divergence between model predictions and given labels, while the dispersion score reflects the model's capacity to handle specific inputs. With accurate labels, these metrics typically align, following an inverse relationship: high loss with low dispersion or low loss with high dispersion \cite{xie2024importance}. The specific combinations of these values correspond to different data characteristics (\textit{e.g.}, rare or common data). However, when this inverse relationship breaks—\textit{i.e.}, a client exhibits unexpectedly high loss relative to its dispersion score—it signals a performance-capacity mismatch, indicating potential label noise. By jointly considering these two metrics, we obtain a more reliable understanding of client data.

	\subsection{Noise-Robust Fair FL Framework}
	With these two metrics reflecting performance and capacity, we then incorporate them into FL via our proposed method, named \textbf{Fed}erated learning via \textbf{P}erformance-\textbf{C}apacity \textbf{A}nalysis (\textbf{FedPCA}). This is designed to robustly ensure fairness by appropriately handling different types of clients. The details are as follows.
	
	\noindent\textbf{Mislabeled Clients Identification.}
	FedPCA begins with client identification using the proposed performance-capacity analysis. In each FL round $t$, client $k$ computes a vector $[\ell_t(\bar{D}_k), S_t(\bar{D}_k)]^\top$, representing its average cross-entropy loss and dispersion score (Eq. \ref{eq:score}) with respect to the global model, characterizing its local dataset. This vector is sent to the server, where a 3-component Gaussian Mixture Model (GMM) is fitted to categorize clients into three sets. Our primary goal is to identify mislabeled clients, which, as discussed earlier, exhibit an unexpectedly high loss relative to their dispersion score. Thus, we designate the group positioned in the upper-right region of the loss-dispersion plane as $\mathcal{S}_n$, representing the set of identified mislabeled clients. This is determined by identifying the centroid that deviates upward and rightward relative to the line connecting the other two centroids.
	The remaining two sets are defined as follows: clients in $\mathcal{S}_c$ exhibit low loss and high dispersion, indicating common data, while clients in $\mathcal{S}_r$ show high loss and low dispersion, likely representing rare data. Visualization results in Fig. \ref{fig:PCA} confirm that this method effectively identifies mislabeled clients.
	
	\noindent\textbf{Selecting Reliable Data for Local Training.}
	To enhance robustness, we construct a new dataset $\hat{D}_{t,k}$ of size $\hat{N}_{t,k}$ from the accessible dataset $\bar{D}_k$, ensuring that client $k$ trains the model on reliable data in round $t$. Specifically, for client $k$ in $\mathcal{S}_c$ and $\mathcal{S}_r$, we use the original dataset for local training, setting $\hat{D}_{t,k} = \bar{D}_k$. However, for client $k$ in $\mathcal{S}_n$, we design two strategies to construct $\hat{D}_{t,k}$. The first strategy, Drop (D), excludes all local data, so $\hat{D}_{t,k} = \O$. The second strategy, High-Confidence Sampling (HS), selects only high-confidence samples from $\bar{D}_k$. Specifically, a sample $\boldsymbol{x}$, along with its predicted class as the label, is included in $\hat{D}_{t,k}$ if the predicted probability on that class exceeds a threshold $\tau$. This threshold $\tau$ is set as the higher of two values: the average of the top $t/T$ quantile of prediction probabilities, calculated individually for correctly classified samples within each client in $\mathcal{S}_c$, or a preset minimum threshold $\tau_{\text{min}}$, where $t$ is the current FL round and $T$ is the total number of rounds. This setup ensures that the model focuses initially on samples with high label accuracy, gradually incorporating more as training progresses, while the preset minimum prevents excessive noise. Once $\hat{D}_{t,k}$ is obtained, we use it to train the local model. 
	
	\begin{algorithm}[!t]
		\caption{Our proposed solution FedPCA}
		\label{alg:FedPCA}
		\textbf{Input}: Number of rounds $T$, number of clients $K$, local datasets $\{\bar{D}_1, \cdots \bar{D}_K\}$.\\
		\textbf{Output}: Final model parameter $\boldsymbol{\theta}_{T+1}$.
		\begin{algorithmic}[1]
			\State Initialize the model parameter $\boldsymbol{\theta}_1$
			\For{$t = 1, \cdots, T$}
			\For{\textit{Client} $k \in [K]$ in parallel}
			\State $\boldsymbol{\theta}_{t, k} \leftarrow \boldsymbol{\theta}_{t}$. \textcolor{blue}{\footnotesize\hfill$\triangleright$ {Download the global model}}
			\State Compute and upload $[\ell_t(\bar{D}_k), S_t(\bar{D}_k)]^\top$.
			\EndFor
			\State  \textit{Server}: Divide clients into $\mathcal{S}_c, \mathcal{S}_r, \mathcal{S}_n$ by GMM.
			\For{\textit{Client} $k \in [K]$ in parallel}
			\State Construct $\hat{D}_{t,k}$.
			\State $\boldsymbol{\theta}_{t,k} \leftarrow \text{Update}(\boldsymbol{\theta}_{t,k}, \hat{D}_{t,k})$. \textcolor{blue}{\footnotesize\hfill$\triangleright$ {Local training}}
			\EndFor
			\State \textit{Server}: Compute $w_{t,k}$ for $k \in [K]$.
			\textcolor{blue}{\footnotesize\hfill$\triangleright$ {Eq. \ref{eq:weights}}}
			\State \textit{Server}: $\boldsymbol{\theta}_{t+1} \leftarrow \sum_{k=1}^{K} w_{t,k} \boldsymbol{\theta}_{t,k}$.
			\textcolor{blue}{\footnotesize\hfill$\triangleright$ {Global aggregation}}
			\EndFor
		\end{algorithmic}
	\end{algorithm}
	
	\noindent\textbf{Integrated Aggregation for Robustness and Fairness.}
	After local training on $\hat{D}_{t,k}$, we aggregate the uploaded models from all clients, taking into account both robustness and fairness. The final aggregation weight for each client $k$ in round $t$ is defined as:
	\begin{equation} \label{eq:weights}
		w_{t,k} = \frac{\hat{N}_{t,k} \cdot r_{t,k} \cdot \exp{(-q \cdot S_t(\bar{D}_k))}}{\sum_{i=1}^{K} \hat{N}_{t,i} \cdot r_{t,i} \cdot \exp{(-q \cdot S_t(\bar{D}_i))}},
	\end{equation}
	where $r_{t,k} \in [0,1]$ indicates label reliability, $q$ is a predefined parameter. The term $\hat{N}_{t,k} \cdot r_{t,k}$ captures the number of reliable labels contributed by $\hat{D}_{t,k}$. For clients in $\mathcal{S}_c$ and $\mathcal{S}_r$, where labels are verified, $\hat{N}_{t,k} = N_k$ and $r_{t,k} = 1$, reflecting full confidence in label accuracy. For clients in $\mathcal{S}_n$, we have $\hat{N}_{t,k} \leq N_k$ and $r_{t,k} \leq 1$, and $r_{t,k}$ is set to the average probability on the pseudo labels. This weighting prioritizes trustworthy data by scaling contributions according to the amount of confident data and label reliability.
	
	Fairness is incorporated through the dispersion score $S_t(\bar{D}_k)$, which reflects the model's capacity to handle the inputs. Lower dispersion, indicating more challenging data, results in higher weights, thereby prioritizing learning from clients with underrepresented data. The parameter $q$ provides additional control, allowing further emphasis on such clients as needed.

	\noindent\textbf{Summary of FedPCA.}
	Each round of FedPCA comprises three key steps: 1) conducting performance-capacity analysis to identify mislabeled clients, 2) selecting reliable data for local training, and 3) aggregating models using the weights defined in Eq. \ref{eq:weights}. The overall pipeline is outlined in Alg. \ref{alg:FedPCA}. Compared to FedAvg \cite{FedAvg}, FedPCA only modifies local data selection and global aggregation weights, remaining simple and easy to implement.
	Please note that methods employing the (D) and (HS) robust strategies are denoted as FedPCA (D) and FedPCA (HS), respectively.

	\begin{table*}[!t]
		\centering
		\renewcommand{\arraystretch}{1.1}
		\caption{Performance comparison on CIFAR-10 (mean (\%) $\pm$ standard deviation (\%)). Bold values indicate the best result.}
		\label{tab:SOTA-CIFAR}
		\resizebox{\textwidth}{!}{
			\begin{tabular}{ll|cccc|cccc|cccc}
				\toprule[2pt]
				\multirow{4}{*}{Category}                                                      & \multirow{4}{*}{Method} & \multicolumn{4}{c|}{$(\rho=0.2, \eta=1.0)$}                                         & \multicolumn{4}{c|}{$(\rho=0.4, \eta=0.8)$}                                         & \multicolumn{4}{c}{$(\rho=0.6, \eta=0.5)$}                                         \\ \cmidrule(lr){3-14}
				&                         & \multicolumn{2}{c}{Worst}                & \multicolumn{2}{c|}{Average}             & \multicolumn{2}{c}{Worst}                & \multicolumn{2}{c|}{Average}             & \multicolumn{2}{c}{Worst}                & \multicolumn{2}{c}{Average}             \\ \cmidrule(lr){3-14}
				&                         & ACC                 & AUC                & ACC                 & AUC                & ACC                 & AUC                & ACC                 & AUC                & ACC                 & AUC                & ACC                & AUC                \\ \midrule[1.2pt]
				Vanilla                                                                        & FedAvg                  & $63.99_{\pm 6.21}$  & $94.73_{\pm 1.36}$ & $74.93_{\pm 5.18}$  & $96.71_{\pm 0.96}$ & $59.03_{\pm 5.13}$  & $94.23_{\pm 1.24}$ & $71.38_{\pm 4.08}$  & $96.23_{\pm 0.95}$ & $54.95_{\pm 5.25}$  & $93.96_{\pm 0.63}$ & $69.50_{\pm 3.23}$ & $96.32_{\pm 0.35}$ \\ \midrule[1.2pt]
				\multirow{4}{*}{Fair FL}                                                       & q-FedAvg                & $65.65_{\pm 3.85}$  & $94.74_{\pm 1.24}$ & $76.28_{\pm 3.66}$  & $96.72_{\pm 0.93}$ & $64.87_{\pm 3.32}$  & $94.89_{\pm 0.99}$ & $74.92_{\pm 2.94}$  & $96.68_{\pm 0.80}$ & $64.17_{\pm 3.64}$  & $93.22_{\pm 1.94}$ & $71.60_{\pm 3.64}$ & $95.05_{\pm 1.56}$ \\ 
				& FedCE                   & $59.80_{\pm 10.06}$ & $92.64_{\pm 3.08}$ & $67.34_{\pm 11.43}$ & $94.67_{\pm 2.65}$ & $48.15_{\pm 14.35}$ & $91.54_{\pm 3.33}$ & $58.74_{\pm 13.36}$ & $93.83_{\pm 2.71}$ & $54.57_{\pm 3.45}$  & $92.66_{\pm 0.72}$ & $67.51_{\pm 2.37}$ & $95.35_{\pm 0.42}$ \\ 
				& FedGA                   & $58.51_{\pm 9.88}$  & $92.77_{\pm 4.23}$ & $73.35_{\pm 8.64}$  & $95.74_{\pm 2.95}$ & $41.24_{\pm 6.94}$  & $85.43_{\pm 3.36}$ & $52.97_{\pm 8.00}$  & $89.70_{\pm 2.91}$ & $57.80_{\pm 2.25}$  & $92.37_{\pm 0.78}$ & $69.09_{\pm 1.49}$ & $95.08_{\pm 0.46}$ \\ 
				& FedISM                  & $69.05_{\pm 5.48}$  & $97.23_{\pm 0.43}$ & $78.45_{\pm 3.15}$  & $98.22_{\pm 0.25}$ & $75.26_{\pm 1.33}$  & $97.72_{\pm 0.11}$ & $80.67_{\pm 1.04}$  & $98.33_{\pm 0.10}$ & $78.35_{\pm 0.76}$  & $97.90_{\pm 0.08}$ & $82.09_{\pm 0.42}$ & $98.41_{\pm 0.04}$ \\ \midrule[1.2pt]
				\multirow{5}{*}{\begin{tabular}[c]{@{}l@{}}Fair FL +\\ Robust FL\end{tabular}} & q-FedAvg + Median       & $63.10_{\pm 5.68}$  & $95.40_{\pm 0.86}$ & $74.54_{\pm 5.21}$  & $97.20_{\pm 0.64}$ & $60.85_{\pm 5.39}$  & $94.65_{\pm 1.13}$ & $71.01_{\pm 5.06}$  & $96.43_{\pm 0.90}$ & $63.87_{\pm 2.10}$  & $94.04_{\pm 0.62}$ & $73.69_{\pm 1.06}$ & $96.09_{\pm 0.35}$ \\ 
				& FedCE + RFA             & $61.39_{\pm 7.48}$  & $94.32_{\pm 1.84}$ & $72.42_{\pm 6.59}$  & $96.33_{\pm 1.35}$ & $53.80_{\pm 7.44}$  & $94.13_{\pm 1.39}$ & $66.80_{\pm 5.75}$  & $96.08_{\pm 1.04}$ & $60.11_{\pm 2.36}$  & $94.38_{\pm 0.38}$ & $72.47_{\pm 1.19}$ & $96.54_{\pm 0.21}$ \\ 
				& FedGA + FedCorr         & $45.07_{\pm 12.29}$ & $92.55_{\pm 1.48}$ & $65.05_{\pm 5.71}$  & $95.42_{\pm 1.01}$ & $41.43_{\pm 10.73}$ & $88.93_{\pm 2.89}$ & $60.29_{\pm 7.41}$  & $93.04_{\pm 1.87}$ & $50.11_{\pm 13.68}$ & $92.40_{\pm 2.50}$ & $66.11_{\pm 7.32}$ & $95.20_{\pm 1.30}$ \\ 
				& FedISM + FedNoRo        & $76.48_{\pm 0.71}$  & $97.84_{\pm 0.09}$ & $83.31_{\pm 0.36}$  & $98.64_{\pm 0.05}$ & $76.17_{\pm 0.87}$  & $97.81_{\pm 0.10}$ & $82.45_{\pm 0.54}$  & $98.57_{\pm 0.05}$ & $78.46_{\pm 0.76}$  & $97.96_{\pm 0.09}$ & $83.03_{\pm 0.55}$ & $98.57_{\pm 0.05}$ \\ 
				& H-nobs (Oracle)                 & $74.55_{\pm 1.34}$  & $97.20_{\pm 0.21}$ & $83.27_{\pm 0.69}$  & $98.39_{\pm 0.11}$ & $77.63_{\pm 1.12}$  & $97.59_{\pm 0.13}$ & $84.13_{\pm 0.57}$  & $98.50_{\pm 0.06}$ & $77.90_{\pm 4.42}$  & $97.45_{\pm 0.73}$ & $81.76_{\pm 1.43}$ & $98.12_{\pm 0.27}$ \\ \midrule[1.2pt]
				\multirow{2}{*}{Ours}                                                          & FedPCA (D)              & $79.78_{\pm 1.07}$  & $\bm{98.22_{\pm 0.12}}$ & $\bm{84.95_{\pm 0.49}}$  & $\bm{98.83_{\pm 0.05}}$ & $81.40_{\pm 0.85}$  & $\bm{98.35_{\pm 0.10}}$ & $\bm{85.20_{\pm 0.21}}$  & $\bm{98.83_{\pm 0.03}}$ & $\bm{81.48_{\pm 0.56}}$  & $\bm{98.28_{\pm 0.08}}$ & $\bm{84.18_{\pm 0.35}}$ & $\bm{98.66_{\pm 0.05}}$ \\ 
				& FedPCA (HS)             & $\bm{79.99_{\pm 1.19}}$  & $\bm{98.22_{\pm 0.12}}$ & $84.88_{\pm 0.41}$  & $98.81_{\pm 0.05}$ & $\bm{81.42_{\pm 0.64}}$  & $98.33_{\pm 0.08}$ & $84.50_{\pm 0.27}$  & $98.77_{\pm 0.03}$ & $80.79_{\pm 0.60}$  & $98.20_{\pm 0.05}$ & $82.38_{\pm 0.45}$ & $98.45_{\pm 0.10}$ \\ \bottomrule[2pt]
			\end{tabular}
		}
	\end{table*}

	\section{Experiment}
	The proposed solution is evaluated on three image classification datasets. We compare it with state-of-the-art methods to demonstrate its effectiveness in addressing the studied problem. Additionally, extensive analytical studies are conducted to gain deeper insights into its performance. More details and results are provided in the supplementary material due to the page limit.

	\begin{table*}[!t]
		\centering
		\renewcommand{\arraystretch}{1.1}
		\caption{Performance comparison on RSNA-ICH (mean (\%) $\pm$ standard deviation (\%)). Bold values indicate the best result.}
		\label{tab:SOTA-ICH}
		\resizebox{\textwidth}{!}{
			\begin{tabular}{ll|cccc|cccc|cccc}
				\toprule[2pt]
				\multirow{4}{*}{Category}                                                      & \multirow{4}{*}{Method} & \multicolumn{4}{c|}{$(\rho=0.1, \eta=1.0)$}                                         & \multicolumn{4}{c|}{$(\rho=0.2, \eta=0.8)$}                                       & \multicolumn{4}{c}{$(\rho=0.4, \eta=0.6)$}                                         \\ \cmidrule(lr){3-14}
				&                         & \multicolumn{2}{c}{Worst}                & \multicolumn{2}{c|}{Average}             & \multicolumn{2}{c}{Worst}               & \multicolumn{2}{c|}{Average}            & \multicolumn{2}{c}{Worst}                & \multicolumn{2}{c}{Average}             \\ \cmidrule(lr){3-14}
				&                         & ACC                 & AUC                & ACC                & AUC                 & ACC                & AUC                & ACC                & AUC                & ACC                & AUC                 & ACC                & AUC                \\ \midrule[1.2pt]

				Vanilla                                                                        & FedAvg                  & $74.41_{\pm 1.09}$  & $84.49_{\pm 0.59}$ & $79.48_{\pm 0.67}$ & $ 88.44_{\pm 0.34}$ & $73.93_{\pm 0.99}$ & $82.87_{\pm 0.94}$ & $79.02_{\pm 0.60}$ & $87.28_{\pm 0.53}$ & $73.07_{\pm 0.84}$ & $80.90_{\pm 0.72}$  & $77.32_{\pm 0.45}$ & $85.04_{\pm 0.36}$ \\ \midrule[1.2pt]
				
				\multirow{4}{*}{Fair FL}                                                       & q-FedAvg                & $75.36_{\pm 1.05}$  & $83.96_{\pm 0.85}$ & $79.85_{\pm 0.74}$ & $87.80_{\pm 0.57}$  & $73.80_{\pm 1.26}$ & $82.69_{\pm 1.13}$ & $78.24_{\pm 1.00}$ & $86.74_{\pm 0.75}$ & $74.20_{\pm 0.90}$ & $81.80_{\pm 0.78}$  & $77.62_{\pm 0.43}$ & $85.06_{\pm 0.42}$ \\ 
				& FedCE                   & $75.91_{\pm 0.82}$  & $83.57_{\pm 0.57}$ & $79.42_{\pm 0.61}$ & $86.93_{\pm 0.51}$  & $74.22_{\pm 0.90}$ & $81.83_{\pm 1.36}$ & $76.66_{\pm 2.14}$ & $85.69_{\pm 0.90}$ & $73.42_{\pm 0.81}$ & $80.79_{\pm 0.79}$  & $76.70_{\pm 0.50}$ & $84.17_{\pm 0.45}$ \\ 
				& FedGA                   & $75.41_{\pm 0.91}$  & $83.71_{\pm 0.70}$ & $78.02_{\pm 0.95}$ & $86.19_{\pm 0.70}$  & $76.82_{\pm 0.92}$ & $84.64_{\pm 1.04}$ & $79.00_{\pm 1.02}$ & $86.93_{\pm 0.86}$ & $76.68_{\pm 0.79}$ & $ 84.53_{\pm 0.64}$ & $79.43_{\pm 0.55}$ & $87.07_{\pm 0.42}$ \\ 
				& FedISM                  & $72.87_{\pm 1.84}$  & $86.39_{\pm 0.44}$ & $75.64_{\pm 1.39}$ & $89.58_{\pm 0.28}$  & $74.51_{\pm 2.63}$ & $86.44_{\pm 0.51}$ & $78.79_{\pm 2.03}$ & $89.76_{\pm 0.35}$ & $78.88_{\pm 0.71}$ & $86.94_{\pm 0.53}$  & $81.72_{\pm 0.27}$ & $89.75_{\pm 0.29}$ \\ \midrule[1.2pt]
				
				\multirow{5}{*}{\begin{tabular}[c]{@{}l@{}}Fair FL +\\ Robust FL\end{tabular}} & q-FedAvg + Median       & $73.00_{\pm 1.71}$  & $83.14_{\pm 0.86}$ & $78.71_{\pm 1.08}$ & $87.63_{\pm 0.44}$  & $71.59_{\pm 1.81}$ & $80.15_{\pm 0.86}$ & $77.45_{\pm 1.00}$ & $85.64_{\pm 0.52}$ & $72.88_{\pm 1.13}$ & $80.23_{\pm 1.18}$  & $77.16_{\pm 0.72}$ & $84.48_{\pm 0.71}$ \\ 
				& FedCE + RFA             & $74.96_{\pm 1.09}$  & $83.47_{\pm 0.49}$ & $79.74_{\pm 0.60}$ & $87.61_{\pm 0.33}$  & $73.67_{\pm 0.82}$ & $82.21_{\pm 0.64}$ & $78.67_{\pm 0.56}$ & $86.60_{\pm 0.49}$ & $72.61_{\pm 0.88}$ & $80.36_{\pm 0.85}$  & $76.79_{\pm 0.48}$ & $84.45_{\pm 0.45}$ \\
				& FedGA + FedCorr         & $75.91_{\pm 1.24}$  & $84.30_{\pm 0.50}$ & $78.38_{\pm 0.81}$ & $87.01_{\pm 0.35}$  & $74.56_{\pm 2.05}$ & $84.55_{\pm 0.74}$ & $78.48_{\pm 0.95}$ & $87.55_{\pm 0.57}$ & $71.17_{\pm 8.79}$ & $82.23_{\pm 3.96}$  & $76.13_{\pm 4.71}$ & $85.74_{\pm 2.32}$ \\ 
				& FedISM + FedNoRo        & $68.80_{\pm 11.73}$ & $85.75_{\pm 1.75}$ & $74.36_{\pm 9.29}$ & $89.48_{\pm 1.03}$  & $72.92_{\pm 7.50}$ & $86.19_{\pm 1.63}$ & $77.90_{\pm 5.21}$ & $89.60_{\pm 0.91}$ & $77.33_{\pm 0.24}$ & $85.48_{\pm 0.33}$  & $80.89_{\pm 0.20}$ & $89.01_{\pm 0.18}$ \\ 
				& H-nobs (Oracle)                 & $77.49_{\pm 0.63}$  & $86.59_{\pm 0.42}$ & $81.25_{\pm 0.35}$ & $89.73_{\pm 0.23}$  & $77.60_{\pm 0.81}$ & $86.82_{\pm 0.36}$ & $81.00_{\pm 0.46}$ & $89.65_{\pm 0.18}$ & $77.16_{\pm 2.53}$ & $86.31_{\pm 1.39}$  & $80.08_{\pm 0.85}$ & $88.81_{\pm 0.49}$ \\ \midrule[1.2pt]
				
				\multirow{2}{*}{Ours}                                                          & FedPCA (D)              & $80.19_{\pm 0.36}$  & $88.93_{\pm 0.14}$ & $83.01_{\pm 0.23}$ & $91.37_{\pm 0.08}$  & $80.13_{\pm 0.28}$ & $88.89_{\pm 0.18}$ & $82.81_{\pm 0.22}$ & $91.22_{\pm 0.10}$ & $79.75_{\pm 0.37}$ & $88.24_{\pm 0.21}$  & $82.26_{\pm 0.25}$ & $90.40_{\pm 0.09}$ \\
				& FedPCA (HS)             & $\bm{80.44_{\pm 0.28}}$  & $\bm{89.06_{\pm 0.16}}$ & $\bm{83.05_{\pm 0.22}}$ & $\bm{91.43_{\pm 0.10}}$  & $\bm{80.38_{\pm 0.37}}$ & $\bm{89.22_{\pm 0.15}}$ & $\bm{82.90_{\pm 0.21}}$ & $\bm{91.38_{\pm 0.08}}$ & $\bm{80.26_{\pm 0.75}}$ & $\bm{89.12_{\pm 0.19}}$  & $\bm{82.43_{\pm 0.38}}$ & $\bm{90.91_{\pm 0.09}}$ \\ \bottomrule[2pt]
			\end{tabular}
		}
	\end{table*}

	\begin{table}[!t]
		\centering
		\renewcommand{\arraystretch}{0.9}
		\caption{Performance comparison on ISIC 2019 (mean (\%) $\pm$ standard deviation (\%)). Bold values indicate the best result.}
		\label{tab:SOTA-ISIC2019}
		\resizebox{\columnwidth}{!}{
			\begin{tabular}{l|cccc}
				\toprule[2pt]
				\multirow{4}{*}{Method} & \multicolumn{4}{c}{$(\rho=0.1, \eta=1.0)$}                                       \\ \cmidrule(lr){2-5}
				& \multicolumn{2}{c}{Worst}               & \multicolumn{2}{c}{Average}            \\ \cmidrule(lr){2-5} 
				& ACC                & AUC                & ACC                & AUC                \\ \midrule[1.2pt]
				FedAvg                  & $37.79_{\pm 4.20}$ & $78.92_{\pm 2.70}$ & $52.80_{\pm 2.99}$ & $86.16_{\pm 1.57}$ \\ \midrule[1.2pt]
				q-FedAvg                & $43.42_{\pm 2.45}$ & $81.86_{\pm 1.11}$ & $54.78_{\pm 1.43}$ & $87.26_{\pm 0.46}$ \\
				FedCE                   & $41.96_{\pm 2.07}$ & $81.69_{\pm 0.87}$ & $53.48_{\pm 1.46}$ & $86.74_{\pm 0.50}$ \\ 
				FedGA                   & $47.97_{\pm 1.23}$ & $84.51_{\pm 0.62}$ & $53.32_{\pm 0.98}$ & $87.09_{\pm 0.60}$ \\ 
				FedISM                  & $37.38_{\pm 6.00}$ & $82.64_{\pm 1.39}$ & $50.41_{\pm 3.31}$ & $87.92_{\pm 0.67}$ \\ \midrule[1.2pt]
				q-FedAvg + Median       & $38.65_{\pm 4.47}$ & $79.09_{\pm 2.53}$ & $54.07_{\pm 2.49}$ & $86.45_{\pm 1.28}$ \\ 
				FedCE + RFA             & $39.00_{\pm 2.83}$ & $79.41_{\pm 1.59}$ & $53.93_{\pm 1.84}$ & $86.42_{\pm 0.86}$ \\ 
				FedGA + FedCorr         & $22.74_{\pm 1.64}$ & $68.97_{\pm 1.50}$ & $45.15_{\pm 1.16}$ & $81.45_{\pm 0.77}$ \\ 
				FedISM + FedNoRo        & $29.21_{\pm 1.68}$ & $79.57_{\pm 0.70}$ & $43.93_{\pm 1.60}$ & $86.44_{\pm 0.41}$ \\ 
				H-nobs (Oracle)                  & $47.41_{\pm 1.18}$ & $84.37_{\pm 0.71}$ & $56.02_{\pm 0.77}$ & $88.39_{\pm 0.43}$ \\ \midrule[1.2pt]
				FedPCA (D)              & $\bm{54.42_{\pm 0.98}}$ & $\bm{88.72_{\pm 0.38}}$ & $59.14_{\pm 0.86}$ & $90.59_{\pm 0.34}$ \\ 
				FedPCA (HS)             & $54.37_{\pm 0.73}$ & $88.70_{\pm 0.31}$ & $\bm{59.22_{\pm 0.92}}$ & $\bm{90.62_{\pm 0.35}}$ \\ \bottomrule[2pt]
			\end{tabular}
		}
	\end{table}
	
	\subsection{Experimental Setup} \label{sec:setup}
	\noindent\textbf{Datasets.}
	To ensure representative evaluation, we select three widely-used classification datasets:
	\begin{itemize}
		\item CIFAR-10 \cite{krizhevsky2009learning}: A benchmark dataset including 10 classes with 50,000 training and 10,000 testing images, widely used in robustness \cite{FedCorr} and fairness \cite{SharpDRO} studies.
		
		\item RSNA ICH \cite{flanders2020construction}: A real-world CT brain image dataset labeled as healthy or one of five intracranial hemorrhage subtypes. Following \cite{jiang2023client}, we randomly select 25,000 images for binary diseased-or-healthy classification.
		
		\item ISIC 2019 \cite{tschandl2018ham10000,codella2018skin,combalia2019bcn20000}: A dataset of skin disease images across eight classes, collected from real-world medical institutions. It exhibits severe class imbalance (the largest-to-smallest class ratio is approximately 54:1), making it a suitable for evaluating method effectiveness under non-ideal conditions.
	\end{itemize}
	For RSNA ICH and ISIC 2019, images are resized to $224 \times 224$ pixels and split into training and testing sets (8:2) following \cite{FedISM}.
	
	\noindent\textbf{Decentralized Data Setting.}
	All training sets are randomly partitioned into multiple subsets—50 for CIFAR-10 and 20 for RSNA ICH and ISIC 2019—to simulate an FL environment. To create common and rare data clients, we corrupt images from minority clients. This approach is widely used to create diverse data distributions (\textit{i.e.}, non-corrupted vs. corrupted images) in both FL \cite{FedISM} and conventional centralized learning \cite{SharpDRO}. Specifically, following \cite{hendrycks2019benchmarking}, we apply Gaussian noise to images from 10\% of clients in CIFAR-10 and 20\% in RSNA ICH and ISIC 2019, resulting in 8:2 and 9:1 data scale imbalances across distributions, respectively. To simulate label noise, we randomly flip a fraction of labels in a subset of common data clients, with the proportions of mislabeled clients and flipped labels controlled by $\rho$ and $\eta$. More general and complex settings are explored in Sec. \ref{sec:appendix_experiments_data_dirtibution} and Sec. \ref{sec:appendix_experiments_noise_dirtibution} (see supplementary material).
	
	\noindent\textbf{Model.}
	ResNet-18 \cite{he2016deep} is adopted as the model for standard evaluation. For CIFAR-10, the model is trained from scratch, while an ImageNet-pretrained model \cite{deng2009imagenet} is used for the other two datasets. 
	
	\noindent\textbf{Implementation Details.}
	For CIFAR-10, each client trains locally with a batch size of 64 using the SGD optimizer with a learning rate of 0.03, momentum of 0.9, and weight decay of 0.0005. Training spans 500 communication rounds, with 1 local epoch per round. For RSNA ICH and ISIC 2019, each client employs a batch size of 32 and trains using the Adam optimizer with a fixed learning rate of 0.0003, beta values (0.9, 0.999), and the same weight decay. Training consists of up to 300 communication rounds for RSNA ICH and 100 for ISIC 2019, with 1 local epoch per round. Logit adjustment \cite{menonlong} is used when training local models on ISIC 2019 to mitigate class imbalance.
	Above-mentioned settings are kept consistent across experiments to ensure a fair comparison.
	For our proposed method, we set $\tau_{\text{min}} = 0.9$ and choose $q$ as 3, 1, and 0.4 for CIFAR-10, RSNA ICH, and ISIC 2019, respectively. To stabilize training, we warm up the model with FedAvg \cite{FedAvg} for 10 rounds before applying our proposed framework. The aggregation weights and client identification results are smoothed over time. Additionally, we integrate sharpness-aware minimization \cite{zhuang2022surrogate}, following recent advances in fair FL research \cite{FedISM}.
	
	\noindent\textbf{Evaluation Strategy.}
	The primary objective is to robustly achieve Rawlsian Max-Min fairness \cite{rawls2001justice}, as formulated in Eq. \ref{eq:objective}. Therefore, we apply the same corruption type to the testing set as was used for subsets of the training clients, generating both corrupted and non-corrupted test sets. The final FL model is evaluated on both to assess its ability to handle both common and rare data, using classification metrics: accuracy (ACC) and the area under the receiver operating characteristic curve (AUC).
	To mitigate random effects, all experiments are repeated three times, with the mean and standard deviation computed over the final five communication rounds, following \cite{huang2023rethinking}.

	\subsection{Comparison to State-of-the-Arts} \label{sec:sota}
	To show superiority of our method, several leading FL methods are selected for comparison, including vanilla FL: FedAvg \pub{AISTATS'17} \cite{FedAvg}; fair FL methods: q-FedAvg \pub{ICLR'20} \cite{q-FedAvg}, FedCE \pub{CVPR'23} \cite{FedCE}, FedGA \pub{CVPR'23} \cite{FedGA}, and FedISM \pub{IJCAI'24} \cite{FedISM}. To mitigate the effects of label noise, we also incorporate robust FL methods—Median \pub{ICML'18} \cite{Median}, RAF \pub{IEEE TSP'22} \cite{RFA}, FedCorr \pub{CVPR'22} \cite{FedCorr}, and FedNoRo \pub{IJCAI'23} \cite{FedNoRo}—in combination with the fair FL methods listed above. Additionaly, we also consider H-nobs \pub{NeurIPS'23} \cite{zhou2024h}, a work that actively combine a robust techique with q-FedAvg. This method requires a parameter specifying the number of clients affected by label poisoning attacks. To maximize its performance, we assume it has prior knowledge of this number, though such information is typically unavailable in real-world scenarios, denoting it as H-nobs (Oracle). In contrast, our method and other baselines operate without this prior knowledge.
	More details of these methods are provided in Sec. \ref{sec:appendix_experiments_detail} of the supplementary material.
	
	Following standard practices for imbalanced scenarios \cite{yang2023change}, we evaluate both the worst-case and average performance across distributions to assess the model's ability to handle diverse data. Extensive experiments are conducted under different noise conditions (\textit{i.e.}, varying $(\rho, \eta)$ pairs) for a comprehensive evaluation. Results are summarized in Tab. \ref{tab:SOTA-CIFAR}, Tab. \ref{tab:SOTA-ICH}, and Tab. \ref{tab:SOTA-ISIC2019}. Additional evaluations on more complex decentralized data settings are included Sec. \ref{sec:appendix_experiments_data_dirtibution} and Sec. \ref{sec:appendix_experiments_noise_dirtibution} in the supplementary materials.
	
	The results reveal three key findings: 1) FedAvg \cite{FedAvg} fails to address this challenge, as it lacks explicit considerations for fairness and robustness. 2) Fair FL methods are vulnerable to label noise, and even when combined with robust techniques, they remain ineffective. While some fair FL methods (\textit{e.g.}, FedISM \cite{FedISM}) improve fairness under label noise, and robust techniques show benefits in specific cases, these effects are inconsistent across settings and cannot be generalized beyond selective favorable cases. 3) Our method consistently achieves the best performance, demonstrating its effectiveness in dealing with this problem regardless of dataset, noise rate, or data partitioning strategy.
	
	\begin{figure}[!t] 
		\centering
		\includegraphics[width=1.0\columnwidth]{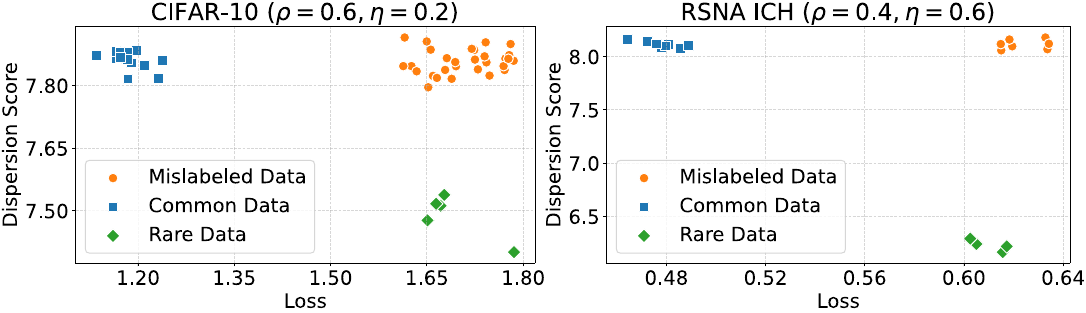}
		\caption{Visualization of performance-capacity analysis.}
		\label{fig:PCA}
	\end{figure}

	\subsection{Analytical Studies}
	
	\noindent\textbf{Visualization of Performance-Capacity Analysis.} 
	We demonstrate how the proposed method effectively differentiates mislabeled clients while preserving those with rare data. Specifically, Fig. \ref{fig:PCA} visualizes clients in the space of the two proposed metrics. The figure shows that both mislabeled and rare data clients exhibit higher loss than common data clients, even overlapping in this metric. As discussed in Sec. \ref{sec:PA}, relying solely on loss leads to trade-offs—either up-weighting high-loss clients, amplifying mislabeled data, or down-weighting them, unfairly penalizing rare data. In contrast, our approach, which jointly considers performance and capacity, reveals a key distinction: mislabeled clients consistently exhibit unexpectedly high loss relative to their dispersion score, positioning them in the upper-right region. This stable separation helps mitigate the competition between fairness and robustness, enabling FedPCA to achieve fairness robustly.
	
	\begin{figure}[!t] 
		\centering
		\includegraphics[width=1.0\columnwidth]{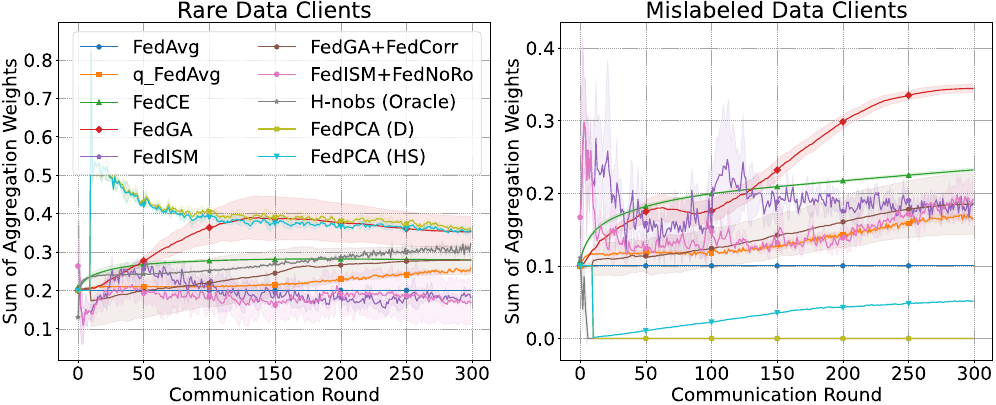}
		\caption{Aggregation weights assigned to clients over communication rounds, with the transparent area indicating the standard deviation. The legend in the first figure remains consistent across both figures. Experiments are conducted on RSNA ICH.}
		\label{fig:weights}
	\end{figure}
	
	\noindent\textbf{Analysis of Focused Clients.} \label{sec:weights}
	To understand how our solution achieves fairness in the presence of noisy label clients, we analyze client prioritization by visualizing the aggregation weights assigned to rare and mislabeled data clients in Fig. \ref{fig:weights}. For comparison, we include weights from other methods discussed in Sec. \ref{sec:sota}, excluding Median \cite{Median} and RFA \cite{RFA} due to their median-based aggregation, which complicates individual client weight quantification. The figures show that our solution achieves two key objectives: 1) Clients with rare data receive consistently higher weights than in other methods, and 2) Clients with noisy labels are either entirely excluded in FedPCA (D) or given minimal weight when selectively incorporating corrected data in FedPCA (HS). This highlights our method's ability to mitigate the trade-off between robustness and fairness, robustly enabling fairness despite label noise. Other methods struggle to achieve both. For example, while FedGA \cite{FedGA} assigns high weights to rare data clients similar to our approach, it still allocates considerable weight to mislabeled clients. When combined with FedCorr \cite{FedCorr}, attention to mislabeled data is reduced, but at the cost of also lowering the weight for rare data. These observations illustrate why our method effectively handles this complex scenario, aligning with its design objectives. More discussion and results can be found in Sec. \ref{sec:appendix_clients_prioritized} in the supplemantary material.

	\begin{table}[!t]
		\centering
		\renewcommand{\arraystretch}{0.9}
		\caption{Results of the component-wise ablation study.}
		\label{tab:ablation}
		\resizebox{1.0\columnwidth}{!}{
			\begin{tabular}{l|l|cccc}
				\toprule[2pt]
				& \multicolumn{1}{c|}{}                         & \multicolumn{2}{c}{Worst}                                                           & \multicolumn{2}{c}{Average}                                                         \\ \cmidrule(lr){3-6} 
				\multirow{-2}{*}{Dataset}                                                                     & \multicolumn{1}{l|}{\multirow{-2}{*}{Method}} & ACC                                      & AUC                                      & ACC                                      & AUC                                      \\ \midrule[1.2pt]
				& w/o Fairness                                  & $67.78_{\pm 0.96}$                         & $96.38_{\pm 0.16}$                         & $79.98_{\pm 0.48}$                         & $98.01_{\pm 0.08}$                         \\
				& w/o Robustness                                & $70.96_{\pm 3.85}$                         & $97.43_{\pm 0.28}$                         & $79.05_{\pm 2.31}$                         & $98.29_{\pm 0.21}$                         \\
				\multirow{-3}{*}{\begin{tabular}[c]{@{}l@{}}CIFAR-10\\ $(\rho=0.2,   \eta=1.0)$\end{tabular}} & \cellcolor[HTML]{E4E4E4}FedPCA (D)           & \cellcolor[HTML]{E4E4E4}$79.78_{\pm 1.07}$ & \cellcolor[HTML]{E4E4E4}$98.22_{\pm 0.12}$ & \cellcolor[HTML]{E4E4E4}$84.95_{\pm 0.49}$ & \cellcolor[HTML]{E4E4E4}$98.83_{\pm 0.05}$ \\ \midrule[1.2pt]
				& w/o Fairness                                  & $79.01_{\pm 0.54}$                         & $87.49_{\pm 0.22}$                         & $82.69_{\pm 0.29}$                         & $90.81_{\pm 0.11}$                         \\
				& w/o Robustness                                & $72.83_{\pm 1.80}$                         & $88.42_{\pm 0.23}$                         & $75.64_{\pm 1.62}$                         & $90.89_{\pm 0.12}$                         \\
				\multirow{-3}{*}{\begin{tabular}[c]{@{}l@{}}RSNA ICH\\ $(\rho=0.1,   \eta=1.0)$\end{tabular}} & \cellcolor[HTML]{E4E4E4}FedPCA (D)           & \cellcolor[HTML]{E4E4E4}$80.19_{\pm 0.36}$ & \cellcolor[HTML]{E4E4E4}$88.93_{\pm 0.14}$ & \cellcolor[HTML]{E4E4E4}$83.01_{\pm 0.23}$ & \cellcolor[HTML]{E4E4E4}$91.37_{\pm 0.08}$ \\ 
				\bottomrule[2pt]
			\end{tabular}
		}
	\end{table}

	\noindent\textbf{Component-wise Studies.}
	We perform component-wise studies to show the effectiveness of each design in our solution. Using FedPCA (D) as an example, we selectively remove its fairness and robustness components.The fairness component is removed by setting $q = 0$ in Eq. \ref{eq:weights}, while the robustness one is excluded by setting $\hat{N}_{t,k} \cdot r_{t,k} = 1.0$ and training on the original dataset $\bar{D}_k$. The results in Tab. \ref{tab:ablation} confirm that each component contributes to achieving fairness in the presence of label noise.
	Additionally, we integrate the robustness component (D) with other fair FL methods. As shown in Tab. \ref{tab:combine}, this combination significantly improves fairness, further validating that: 1) Fair FL is affected by label noise, and 2) Our proposed method enhances the robustness of fair FL.

	\begin{table}[!t]
		\centering
		\renewcommand{\arraystretch}{0.82}
		\caption{Performance comparison of fair FL methods and their combinations with our proposed robust component (D).}
		\label{tab:combine}
		\resizebox{1.0\columnwidth}{!}{
			\begin{tabular}{l|l|cccc}
				\toprule[2pt]
				\multirow{2}{*}{Dataset}                                                                   & \multirow{2}{*}{Method}        & \multicolumn{2}{c}{Worst}            & \multicolumn{2}{c}{Average}          \\  \cmidrule(lr){3-6} 
				&                                & ACC               & AUC              & ACC               & AUC              \\ \midrule[1.2pt]
				\multirow{8}{*}{\begin{tabular}[c]{@{}l@{}}CIFAR-10\\ $(\rho=0.2, \eta=1.0)$\end{tabular}} & q-FedAvg       & $65.65_{\pm 3.85}$  & $94.74_{\pm 1.24}$ & $76.28_{\pm 3.66}$  & $96.72_{\pm 0.93}$ \\
				& q-FedAvg  + (D) & $74.12_{\pm 1.53}$  & $97.08_{\pm 0.26}$ & $83.13_{\pm 0.76}$  & $98.33_{\pm 0.13}$ \\ \cmidrule(lr){2-6} 
				& FedCE          & $59.80_{\pm 10.06}$ & $92.64_{\pm 3.08}$ & $67.34_{\pm 11.43}$ & $94.67_{\pm 2.65}$ \\
				& FedCE  + (D)       & $75.79_{\pm 1.22}$  & $97.40_{\pm 0.19}$ & $83.94_{\pm 0.58}$  & $98.49_{\pm 0.09}$ \\ \cmidrule(lr){2-6} 
				& FedGA     & $58.51_{\pm 9.88}$  & $92.77_{\pm 4.23}$ & $73.35_{\pm 8.64}$  & $95.74_{\pm 2.95}$ \\
				& FedGA  + (D)       & $71.24_{\pm 1.22}$  & $96.64_{\pm 0.21}$ & $81.74_{\pm 0.55}$  & $98.12_{\pm 0.11}$ \\ \cmidrule(lr){2-6} 
				& FedISM           & $69.05_{\pm 5.48}$  & $97.23_{\pm 0.43}$ & $78.45_{\pm 3.15}$  & $98.22_{\pm 0.25}$ \\
				& FedISM + (D)     & $81.74_{\pm 0.91}$  & $98.41_{\pm 0.12}$ & $85.57_{\pm 0.35}$  & $98.88_{\pm 0.05}$ \\ \midrule[1.2pt]
				\multirow{8}{*}{\begin{tabular}[c]{@{}l@{}}RSNA ICH\\ $(\rho=0.1, \eta=1.0)$\end{tabular}} & q-FedAvg      & $75.36_{\pm 1.05}$  & $83.96_{\pm 0.85}$ & $79.85_{\pm 0.74}$  & $87.80_{\pm 0.57}$ \\
				& q-FedAvg  + (D) & $76.98_{\pm 0.85}$  & $86.19_{\pm 0.55}$ & $81.00_{\pm 0.43}$  & $89.51_{\pm 0.25}$ \\ \cmidrule(lr){2-6} 
				& FedCE             & $75.91_{\pm 0.82}$  & $83.57_{\pm 0.57}$ & $79.42_{\pm 0.61}$  & $86.93_{\pm 0.51}$ \\
				& FedCE  + (D)       & $77.61_{\pm 0.75}$  & $86.85_{\pm 0.32}$ & $81.27_{\pm 0.45}$  & $89.77_{\pm 0.18}$ \\ \cmidrule(lr){2-6} 
				& FedGA            & $75.41_{\pm 0.91}$  & $83.71_{\pm 0.70}$ & $78.02_{\pm 0.95}$  & $86.19_{\pm 0.70}$ \\
				& FedGA + (D)       & $78.61_{\pm 0.69}$  & $87.19_{\pm 0.27}$ & $81.58_{\pm 0.40}$  & $89.87_{\pm 0.14}$ \\ \cmidrule(lr){2-6} 
				& FedISM          & $72.87_{\pm 1.84}$  & $86.39_{\pm 0.44}$ & $75.64_{\pm 1.39}$  & $89.58_{\pm 0.28}$ \\
				& FedISM  + (D)     & $80.13_{\pm 0.35}$  & $88.62_{\pm 0.27}$ & $82.81_{\pm 0.20}$  & $91.15_{\pm 0.09}$ \\ 
				\bottomrule[2pt]
			\end{tabular}
		}
	\end{table}

	\section{Conclusion}
	In this paper, we tackle the challenge of ensuring performance fairness in FL while maintaining robustness against mislabeled data. While prior research suggests an inherent trade-off between fairness and robustness, we argue that both can be improved by precisely identifying mislabeled clients while preserving rare data clients. To this end, we propose performance-capacity analysis, leveraging two complementary metrics to detect mislabeled clients based on the discrepancy between model performance and its capacity to handle the data. This enables the development of robust and fair strategies that mitigate the impact of mislabeled data while ensuring equitable learning across clients. Extensive experiments validate the effectiveness of our method. We believe this study provides valuable insights for fostering trustworthy FL, particularly in enhancing fairness and robustness.

	{
		\small
		\bibliographystyle{ieeenat_fullname}
		\bibliography{main}

\begin{thebibliography}{45}
\providecommand{\natexlab}[1]{#1}
\providecommand{\url}[1]{\texttt{#1}}
\expandafter\ifx\csname urlstyle\endcsname\relax
  \providecommand{\doi}[1]{doi: #1}\else
  \providecommand{\doi}{doi: \begingroup \urlstyle{rm}\Url}\fi

\bibitem[Chen et~al.(2020)Chen, Kornblith, Norouzi, and Hinton]{chen2020simple}
Ting Chen, Simon Kornblith, Mohammad Norouzi, and Geoffrey Hinton.
\newblock A simple framework for contrastive learning of visual
  representations.
\newblock In \emph{ICML}, pages 1597--1607, 2020.

\bibitem[Codella et~al.(2018)]{codella2018skin}
Noel~CF Codella et~al.
\newblock Skin lesion analysis toward melanoma detection: A challenge at the
  2017 international symposium on biomedical imaging ({ISBI}), hosted by the
  international skin imaging collaboration ({ISIC}).
\newblock In \emph{ISBI}, pages 168--172, 2018.

\bibitem[Combalia et~al.(2019)Combalia, Codella, Rotemberg, Helba, Vilaplana,
  Reiter, Carrera, Barreiro, Halpern, Puig, et~al.]{combalia2019bcn20000}
Marc Combalia, Noel~CF Codella, Veronica Rotemberg, Brian Helba, Veronica
  Vilaplana, Ofer Reiter, Cristina Carrera, Alicia Barreiro, Allan~C Halpern,
  Susana Puig, et~al.
\newblock {BCN20000}: Dermoscopic lesions in the wild.
\newblock \emph{arXiv:1908.02288}, 2019.

\bibitem[Deng et~al.(2009)Deng, Dong, Socher, Li, Li, and
  Fei-Fei]{deng2009imagenet}
Jia Deng, Wei Dong, Richard Socher, Li-Jia Li, Kai Li, and Li Fei-Fei.
\newblock Imagenet: A large-scale hierarchical image database.
\newblock In \emph{CVPR}, 2009.

\bibitem[Devijver and Kittler(1982)]{devijver1982pattern}
Pierre~A Devijver and Josef Kittler.
\newblock \emph{Pattern recognition: A statistical approach}.
\newblock Prentice hall, 1982.

\bibitem[Dou et~al.(2021)Dou, So, Jiang, Liu, Vardhanabhuti, Kaissis, Li, Si,
  Lee, Yu, et~al.]{dou2021federated}
Qi Dou, Tiffany~Y So, Meirui Jiang, Quande Liu, Varut Vardhanabhuti, Georgios
  Kaissis, Zeju Li, Weixin Si, Heather~HC Lee, Kevin Yu, et~al.
\newblock Federated deep learning for detecting {COVID-19} lung abnormalities
  in {CT:} a privacy-preserving multinational validation study.
\newblock \emph{NPJ Digit. Medicine}, 2021.

\bibitem[Flanders et~al.(2020)Flanders, Prevedello, Shih, Halabi,
  Kalpathy-Cramer, Ball, Mongan, Stein, Kitamura, Lungren,
  et~al.]{flanders2020construction}
Adam~E Flanders, Luciano~M Prevedello, George Shih, Safwan~S Halabi, Jayashree
  Kalpathy-Cramer, Robyn Ball, John~T Mongan, Anouk Stein, Felipe~C Kitamura,
  Matthew~P Lungren, et~al.
\newblock Construction of a machine learning dataset through collaboration: The
  {RSNA} 2019 brain {CT} hemorrhage challenge.
\newblock \emph{Radiology: Artificial Intelligence}, 2\penalty0 (3), 2020.

\bibitem[Garber and Djouadi(1988)]{garber1988bounds}
Frederick~D Garber and Abdelhamid Djouadi.
\newblock Bounds on the bayes classification error based on pairwise risk
  functions.
\newblock \emph{IEEE TPAMI}, 10\penalty0 (2):\penalty0 281--288, 1988.

\bibitem[Han et~al.(2018)Han, Yao, Yu, Niu, Xu, Hu, Tsang, and
  Sugiyama]{han2018co}
Bo Han, Quanming Yao, Xingrui Yu, Gang Niu, Miao Xu, Weihua Hu, Ivor Tsang, and
  Masashi Sugiyama.
\newblock Co-teaching: Robust training of deep neural networks with extremely
  noisy labels.
\newblock In \emph{NeurIPS}, 2018.

\bibitem[He and Garcia(2009)]{he2009learning}
Haibo He and Edwardo~A Garcia.
\newblock Learning from imbalanced data.
\newblock \emph{IEEE TKDE}, 21\penalty0 (9):\penalty0 1263--1284, 2009.

\bibitem[He et~al.(2016)He, Zhang, Ren, and Sun]{he2016deep}
Kaiming He, Xiangyu Zhang, Shaoqing Ren, and Jian Sun.
\newblock Deep residual learning for image recognition.
\newblock In \emph{CVPR}, 2016.

\bibitem[Hendrycks and Dietterich(2019)]{hendrycks2019benchmarking}
Dan Hendrycks and Thomas Dietterich.
\newblock Benchmarking neural network robustness to common corruptions and
  perturbations.
\newblock In \emph{ICLR}, 2019.

\bibitem[Huang et~al.(2023{\natexlab{a}})Huang, Ye, Shi, Li, and
  Du]{huang2023rethinking}
Wenke Huang, Mang Ye, Zekun Shi, He Li, and Bo Du.
\newblock Rethinking federated learning with domain shift: A prototype view.
\newblock In \emph{CVPR}, 2023{\natexlab{a}}.

\bibitem[Huang et~al.(2024)Huang, Ye, Shi, Wan, Li, Du, and
  Yang]{huang2024federated}
Wenke Huang, Mang Ye, Zekun Shi, Guancheng Wan, He Li, Bo Du, and Qiang Yang.
\newblock Federated learning for generalization, robustness, fairness: A survey
  and benchmark.
\newblock \emph{IEEE TPAMI}, 2024.

\bibitem[Huang et~al.(2023{\natexlab{b}})Huang, Zhu, Xia, Shen, Yu, Gong, Han,
  Du, and Liu]{SharpDRO}
Zhuo Huang, Miaoxi Zhu, Xiaobo Xia, Li Shen, Jun Yu, Chen Gong, Bo Han, Bo Du,
  and Tongliang Liu.
\newblock Robust generalization against photon-limited corruptions via
  worst-case sharpness minimization.
\newblock In \emph{CVPR}, 2023{\natexlab{b}}.

\bibitem[Imteaj and Amini(2022)]{imteaj2022leveraging}
Ahmed Imteaj and M~Hadi Amini.
\newblock Leveraging asynchronous federated learning to predict customers
  financial distress.
\newblock \emph{Intell. Syst. Appl.}, 14:\penalty0 200064, 2022.

\bibitem[Jiang et~al.(2023{\natexlab{a}})Jiang, Roth, Li, Yang, Zhao, Nath, Xu,
  Dou, and Xu]{FedCE}
Meirui Jiang, Holger~R Roth, Wenqi Li, Dong Yang, Can Zhao, Vishwesh Nath,
  Daguang Xu, Qi Dou, and Ziyue Xu.
\newblock Fair federated medical image segmentation via client contribution
  estimation.
\newblock In \emph{CVPR}, pages 16302--16311, 2023{\natexlab{a}}.

\bibitem[Jiang et~al.(2023{\natexlab{b}})Jiang, Zhong, Le, Li, and
  Dou]{jiang2023client}
Meirui Jiang, Yuan Zhong, Anjie Le, Xiaoxiao Li, and Qi Dou.
\newblock Client-level differential privacy via adaptive intermediary in
  federated medical imaging.
\newblock In \emph{MICCAI}, pages 500--510. Springer, 2023{\natexlab{b}}.

\bibitem[Khosla et~al.(2020)Khosla, Teterwak, Wang, Sarna, Tian, Isola,
  Maschinot, Liu, and Krishnan]{khosla2020supervised}
Prannay Khosla, Piotr Teterwak, Chen Wang, Aaron Sarna, Yonglong Tian, Phillip
  Isola, Aaron Maschinot, Ce Liu, and Dilip Krishnan.
\newblock Supervised contrastive learning.
\newblock In \emph{NeurIPS}, pages 18661--18673, 2020.

\bibitem[Krizhevsky and Hinton(2009)]{krizhevsky2009learning}
A. Krizhevsky and G. Hinton.
\newblock Learning multiple layers of features from tiny images.
\newblock \emph{Master's thesis, University of Toronto}, 2009.

\bibitem[Li et~al.(2024)Li, Li, Cheng, Liao, and Yu]{li2024feddiv}
Jichang Li, Guanbin Li, Hui Cheng, Zicheng Liao, and Yizhou Yu.
\newblock Feddiv: Collaborative noise filtering for federated learning with
  noisy labels.
\newblock In \emph{AAAI}, 2024.

\bibitem[Li et~al.(2020)Li, Sanjabi, Beirami, and Smith]{q-FedAvg}
Tian Li, Maziar Sanjabi, Ahmad Beirami, and Virginia Smith.
\newblock Fair resource allocation in federated learning.
\newblock In \emph{ICLR}, 2020.

\bibitem[Li et~al.(2021)Li, Hu, Beirami, and Smith]{li2021ditto}
Tian Li, Shengyuan Hu, Ahmad Beirami, and Virginia Smith.
\newblock Ditto: Fair and robust federated learning through personalization.
\newblock In \emph{ICML}, pages 6357--6368. PMLR, 2021.

\bibitem[Liu et~al.(2020)Liu, Niles-Weed, Razavian, and
  Fernandez-Granda]{liu2020early}
Sheng Liu, Jonathan Niles-Weed, Narges Razavian, and Carlos Fernandez-Granda.
\newblock Early-learning regularization prevents memorization of noisy labels.
\newblock In \emph{NeurIPS}, pages 20331--20342, 2020.

\bibitem[Liu et~al.(2022)Liu, Liu, Zhu, Shen, and
  Fernandez-Granda]{liu2022adaptive}
Sheng Liu, Kangning Liu, Weicheng Zhu, Yiqiu Shen, and Carlos Fernandez-Granda.
\newblock Adaptive early-learning correction for segmentation from noisy
  annotations.
\newblock In \emph{CVPR}, pages 2606--2616, 2022.

\bibitem[McMahan et~al.(2017)McMahan, Moore, Ramage, Hampson, and
  y~Arcas]{FedAvg}
Brendan McMahan, Eider Moore, Daniel Ramage, Seth Hampson, and Blaise~Aguera y
  Arcas.
\newblock Communication-efficient learning of deep networks from decentralized
  data.
\newblock In \emph{AISTATS}, pages 1273--1282, 2017.

\bibitem[Menon et~al.()Menon, Jayasumana, Rawat, Jain, Veit, and
  Kumar]{menonlong}
Aditya~Krishna Menon, Sadeep Jayasumana, Ankit~Singh Rawat, Himanshu Jain,
  Andreas Veit, and Sanjiv Kumar.
\newblock Long-tail learning via logit adjustment.
\newblock In \emph{ICLR}.

\bibitem[Mills et~al.(2019)Mills, Hu, and Min]{mills2019communication}
Jed Mills, Jia Hu, and Geyong Min.
\newblock Communication-efficient federated learning for wireless edge
  intelligence in iot.
\newblock \emph{IEEE IoTJ}, 7\penalty0 (7):\penalty0 5986--5994, 2019.

\bibitem[Mohri et~al.(2019)Mohri, Sivek, and Suresh]{mohri2019agnostic}
Mehryar Mohri, Gary Sivek, and Ananda~Theertha Suresh.
\newblock Agnostic federated learning.
\newblock In \emph{ICML}, pages 4615--4625, 2019.

\bibitem[Oord et~al.(2018)Oord, Li, and Vinyals]{oord2018representation}
Aaron van~den Oord, Yazhe Li, and Oriol Vinyals.
\newblock Representation learning with contrastive predictive coding.
\newblock \emph{arXiv preprint arXiv:1807.03748}, 2018.

\bibitem[Pillutla et~al.(2022)Pillutla, Kakade, and Harchaoui]{RFA}
Krishna Pillutla, Sham~M. Kakade, and Zaid Harchaoui.
\newblock {Robust Aggregation for Federated Learning}.
\newblock \emph{IEEE TSP}, 70:\penalty0 1142--1154, 2022.

\bibitem[Rawls(2001)]{rawls2001justice}
John Rawls.
\newblock Justice as fairness: A restatement.
\newblock \emph{Erin Kelly/Harvard University}, 2001.

\bibitem[Tschandl et~al.(2018)Tschandl, Rosendahl, and
  Kittler]{tschandl2018ham10000}
Philipp Tschandl, Cliff Rosendahl, and Harald Kittler.
\newblock The {HAM}10000 dataset, a large collection of multi-source
  dermatoscopic images of common pigmented skin lesions.
\newblock \emph{Scientific Data}, 5\penalty0 (1):\penalty0 1--9, 2018.

\bibitem[Tumer and Ghosh(2003)]{tumer2003bayes}
Kagan Tumer and Joydeep Ghosh.
\newblock Bayes error rate estimation using classifier ensembles.
\newblock \emph{International Journal of Smart Engineering System Design},
  5\penalty0 (2):\penalty0 95--109, 2003.

\bibitem[Wang et~al.(2020)Wang, Sreenivasan, Rajput, Vishwakarma, Agarwal,
  Sohn, Lee, and Papailiopoulos]{wang2020attack}
Hongyi Wang, Kartik Sreenivasan, Shashank Rajput, Harit Vishwakarma, Saurabh
  Agarwal, Jy-yong Sohn, Kangwook Lee, and Dimitris Papailiopoulos.
\newblock Attack of the tails: Yes, you really can backdoor federated learning.
\newblock In \emph{NeurIPS}, pages 16070--16084, 2020.

\bibitem[Wu et~al.(2023)Wu, Yu, Jiang, Cheng, and Yan]{FedNoRo}
Nannan Wu, Li Yu, Xuefeng Jiang, Kwang{-}Ting Cheng, and Zengqiang Yan.
\newblock Fednoro: Towards noise-robust federated learning by addressing class
  imbalance and label noise heterogeneity.
\newblock In \emph{IJCAI}, 2023.

\bibitem[Wu et~al.(2024)Wu, Kuang, Yan, and Yu]{FedISM}
Nannan Wu, Zhuo Kuang, Zengqiang Yan, and Li Yu.
\newblock From optimization to generalization: Fair federated learning against
  quality shift via inter-client sharpness matching.
\newblock In \emph{IJCAI}, 2024.

\bibitem[Xie et~al.(2024)Xie, Wei, Feng, Cao, and An]{xie2024importance}
Renchunzi Xie, Hongxin Wei, Lei Feng, Yuzhou Cao, and Bo An.
\newblock On the importance of feature separability in predicting
  out-of-distribution error.
\newblock In \emph{NeurIPS}, 2024.

\bibitem[Xu et~al.(2022)Xu, Chen, Quek, and Chong]{FedCorr}
Jingyi Xu, Zihan Chen, Tony~QS Quek, and Kai Fong~Ernest Chong.
\newblock {FedCorr}: Multi-stage federated learning for label noise correction.
\newblock In \emph{CVPR}, pages 10184--10193, 2022.

\bibitem[Xu and Lyu(2020)]{xu2020reputation}
Xinyi Xu and Lingjuan Lyu.
\newblock A reputation mechanism is all you need: Collaborative fairness and
  adversarial robustness in federated learning.
\newblock \emph{arXiv preprint arXiv:2011.10464}, 2020.

\bibitem[Yang et~al.(2023)Yang, Zhang, Katabi, and Ghassemi]{yang2023change}
Yuzhe Yang, Haoran Zhang, Dina Katabi, and Marzyeh Ghassemi.
\newblock Change is hard: A closer look at subpopulation shift.
\newblock In \emph{ICML}, 2023.

\bibitem[Yin et~al.(2018)Yin, Chen, Kannan, and Bartlett]{Median}
Dong Yin, Yudong Chen, Ramchandran Kannan, and Peter Bartlett.
\newblock Byzantine-robust distributed learning: Towards optimal statistical
  rates.
\newblock In \emph{ICML}, pages 5650--5659, 2018.

\bibitem[Zhang et~al.(2023)Zhang, Xu, Yao, Zhang, Tian, and Wang]{FedGA}
Ruipeng Zhang, Qinwei Xu, Jiangchao Yao, Ya Zhang, Qi Tian, and Yanfeng Wang.
\newblock Federated domain generalization with generalization adjustment.
\newblock In \emph{CVPR}, 2023.

\bibitem[Zhou et~al.(2023)Zhou, Xu, Wang, and Tian]{zhou2024h}
Guanqiang Zhou, Ping Xu, Yue Wang, and Zhi Tian.
\newblock H-nobs: achieving certified fairness and robustness in distributed
  learning on heterogeneous datasets.
\newblock In \emph{NeurIPS}, 2023.

\bibitem[Zhuang et~al.(2022)Zhuang, Gong, Yuan, Cui, Adam, Dvornek, Tatikonda,
  Duncan, and Liu]{zhuang2022surrogate}
Juntang Zhuang, Boqing Gong, Liangzhe Yuan, Yin Cui, Hartwig Adam, Nicha
  Dvornek, Sekhar Tatikonda, James Duncan, and Ting Liu.
\newblock Surrogate gap minimization improves sharpness-aware training.
\newblock In \emph{ICLR}, 2022.

\end{thebibliography}
	}
	

	\clearpage
	\setcounter{page}{1}
	\appendix
	
	\onecolumn
	\begin{center}
		\Large \textbf{FedPCA: Noise-Robust Fair Federated Learning via Performance-Capacity Analysis} \\
		\vspace{0.3cm}
		\Large {Supplementary Material}
	\end{center}

	\section{Outline}
	This supplementary material provides additional details to enhance the understanding of our work. The content is organized as follows:

	\begin{itemize}[left=0cm] 
		\item Section \ref{sec:appendix_notation} summarizes the mathematical notations used throughout the paper.
		
		\item Section \ref{sec:appendix_experiments_detail} describes the methods used for comparison.
		
		\item Section \ref{sec:appendix_experiments} provides further details and additional results from our experiments:

		\begin{itemize}[label=-, left=0.3cm] 
			\item Section \ref{sec:appendix_experiments_std} evaluates fairness using an alternative metric, \textit{i.e.}, performance uniformity.
			
			\item Section \ref{sec:appendix_clients_prioritized} comprehensively analyzes which clients are prioritized by different methods.
			
			\item Section \ref{sec:appendix_experiments_discussion} analyzes the impact of key hyperparameters in our solution.
			
			\item Section \ref{sec:appendix_experiments_visualization} visualizes the feature space to further illustrate our proposed performance-capacity analysis.
			
			 \item Section \ref{sec:appendix_experiments_data_dirtibution} presents additional comparison results under two more general client data distribution settings.
			
			\item Section \ref{sec:appendix_experiments_noise_dirtibution} extends the comparison to a more general setting concerning label noise distribution.
		\end{itemize}
	\end{itemize}

	\section{Notation Table} \label{sec:appendix_notation}
	To facilitate easier reference and comprehension, we provide a notation table (Tab. \ref{tab:notation}) summarizing the key symbols and variables used throughout the paper.
	
	\begin{table*}[!h]
		\centering
		\caption{Mathematical notations used in this paper.}
		\label{tab:notation}
		\resizebox{1.0\textwidth}{!}{
			\begin{tabular}{ll|ll}
				\toprule[2pt]
				\textbf{Notations}    & \textbf{Description}                                   & \textbf{Notations}                            & \textbf{Description}                                      \\
				\midrule[1.2pt]
				$K$                   & Number of clients                                      & $P(y=i)$                                      & Class prior                                                          \\
				$\mathcal{X}$         & Input space                                            & $\delta$                                      & Feature distance between classes                                     \\
				$\mathcal{Y}$         & Label space                                            & $E_{\text{bayes}}^C$                          & Bayes error rate for multi-class settings                            \\
				$C$                   & Number of classes                                      & $S(\bar{D}_k)$                                & Feature dispersion score of client $k$                               \\
				$D_k$                 & Inaccessible dataset of client $k$ with correct labels & $n_{k,j}$                                     & Sample count for cluster $j$ in client $k$                           \\
				$\boldsymbol x$       & Input                                                  & $\tilde{\boldsymbol{\mu}}_{k,j}$              & Feature centroid for the cluster                                     \\
				$y$                   & Correct label                                          & $\bar{\boldsymbol{\mu}}_k$                    & Mean of all features                                                 \\
				$N_k$                 & Size of $D_k$                                          & $t$                                           & Current round number                                                 \\
				$\mathcal{P}_1$       & Common data distribution                               & $T$                                           & Total communication rounds                                           \\
				$\mathcal{P}_2$       & Rare data distribution                                 & $\ell_t(\bar{D}_k)$                           & Average cross-entropy loss on client $k$ in round $t$                \\
				$\bar{D}$             & Accessible dataset of client $k$ with noisy labels     & $\mathcal{S}_c, \mathcal{S}_r, \mathcal{S}_n$ & Three groups of clients                                              \\
				$\bar{y}_k$           & Noisy label                                            & $\hat{D}_{t,k}$                               & Constructed reliable dataset in client $k$ of round $t$              \\
				$f$                   & Feature extractor                                      & $\hat{N}_{t,k}$                               & Size of $\hat{D}_{t,k}$                                              \\
				$g$                   & Classifier                                             & $\tau$                                        & Threshold for selecting reliable data                                \\
				$g \circ f$           & The entire neural network                              & $\tau_{\text{min}}$                           & Minimum threshold                                                    \\
				$\boldsymbol{\theta}$ & Parameters of the neural network                       & $w_{t,k}$                                     & Aggregation weights of FedPCA for client $k$ in round $t$            \\
				$\mathcal{Z}$         & Feature space                                          & $r_{t,k}$                                     & Average label confidence for $\hat{D}_{t,k}$                         \\
				$\Delta$              & Probability simplex                                    & $q$                                           & Hyper-parameter in $w_{t,k}$                                         \\
				$\ell$                & Loss function                                          & $\rho$                                        & Proportion of clients with incorrect labels                          \\
				$\mathbb{R_+}$        & Set of positive real numbers                           & $\eta$                                        & Proportion of randomly flipped labels in each mislabeled data client \\
				$E_{\text{bayes}}$    & Bayes error rate for binary settings                   &                                               &                                                                      \\
				\bottomrule[2pt]
			\end{tabular}
		}
	\end{table*}

	\section{Details of Comparison Methods}  \label{sec:appendix_experiments_detail}
	
	This section provides an overview of the methods compared in Sec. \ref{sec:sota}, along with their implementation approaches.
	
	\begin{itemize}
		\item \textbf{q-FedAvg} \cite{q-FedAvg}: Designed to address performance fairness, q-FedAvg incorporates client training loss into the global update stage. Following \cite{FedISM}, its implementation is simplified by using a scaling constant to adjust global aggregation, aligning it with FedAvg \cite{FedAvg} via loss-aware weights. The hyper-parameter $q$ is selected from \{0.1, 0.25, 0.5, 1.0, 2.0, 5.0\}, and the best-performing configuration is reported for main results.
		
		\item \textbf{FedCE} \cite{FedCE}: Originally developed to enhance fairness in FL for medical image segmentation, FedCE is adapted in this study to support image classification tasks.
		
		\item \textbf{FedGA} \cite{FedGA}: FedGA focuses on improving domain generalization by promoting fairness across diverse client data distributions. In this study, it is used as a fair FL approach.
		
		\item \textbf{FedISM} \cite{FedISM}: To achieve fairness among client sub-distributions, FedISM maintains uniformly low sharpness across all clients, facilitating fair generalization.
		
		\item \textbf{q-FedAvg} \cite{q-FedAvg} + \textbf{Median} \cite{Median}: The Median method uses a median-based aggregation strategy that minimizes the influence of potentially harmful clients on the global model. We aggregate the models produced by q-FedAvg and Median to benefit from this robust strategy.
		
		\item \textbf{FedCE} \cite{FedCE} + \textbf{RFA} \cite{RFA}: RFA utilizes geometric median-based aggregation for enhanced robustness. By merging models from FedCE and RFA, we achieve a combination that leverages both fairness and robustness.
		
		\item \textbf{FedGA} \cite{FedGA} + \textbf{FedCorr} \cite{FedCorr}: FedCorr identifies noisy clients and incorrect labels through local intrinsic dimensions and loss metrics, followed by label correction to enhance training data reliability. FedCorr operates in three stages (set to 10, 240, and 250 for CIFAR-10 and 10, 140, and 150 for both RSNA ICH and ISIC 2019). We aggregate FedGA and FedCorr models to benefit from both methods.
		
		\item \textbf{FedISM} \cite{FedISM} + \textbf{FedNoRo} \cite{FedNoRo}: FedNoRo detects noisy clients based on per-class loss analysis and employs knowledge distillation and robust aggregation to mitigate the impact of label noise. We aggregate FedISM and FedNoRo models with an additional aggregation to obtain benefits from both approaches.
		
		\item \textbf{H-nobs} \cite{zhou2024h}: H-nobs is a method that integrates a robust mechanism with a fairness-oriented FL approach (\textit{i.e.}, q-FedAvg). It requires a parameter specifying the number of attacked clients (\textit{i.e.}, those with incorrect labels in this paper). The method filters out this number of clients based on the highest gradient norms before applying q-FedAvg to enhance fairness. To maximize its performance, we assume it has prior knowledge of the exact number of attacked clients and sets the parameter accordingly. Ideally, this allows it to precisely identify mislabeled clients while retaining all others for fair learning. We denote this variant as H-nobs (Oracle).
	\end{itemize}

	\section{Additional Experimental Results} \label{sec:appendix_experiments}

	\subsection{Comparison on Performance Uniformity} \label{sec:appendix_experiments_std}
	In Tabs. \ref{tab:SOTA-CIFAR}, \ref{tab:SOTA-ICH}, and \ref{tab:SOTA-ISIC2019}, we compare our method with related approaches based on absolute performance, focusing on both worst-case and average scores across distributions. To complement this evaluation, we also consider performance uniformity using the standard deviation (STD) of ACC and AUC across common and rare data distributions. This metric is widely used to assess performance fairness and serves as a supplementary indicator for evaluating our objective \cite{q-FedAvg}. The results in Tab. \ref{tab:appendix_STD} show that our method generally achieves lower STD values. Combined with the previously discussed performance results, these findings demonstrate that our approach not only maintains high accuracy but also ensures uniform performance across distributions, effectively enhancing fairness in the presence of mislabeled data.

	\begin{table*}[!h]
		\centering
		\renewcommand{\arraystretch}{1.1}
		\caption{Standard deviation ($\downarrow$) comparison across three datasets, with mean (\%) $\pm$ standard deviation (\%) derived from multiple independently conducted experiments with various models. Bold values highlight the best result.}
		\label{tab:appendix_STD}
		\resizebox{\textwidth}{!}{
			\begin{tabular}{ll|cccccc|cccccc|cc}
				\toprule[2pt]
				\multirow{4}{*}{Category}                                                      & \multirow{4}{*}{Method} & \multicolumn{6}{c|}{CIFAR-10}                                                                                                           & \multicolumn{6}{c|}{RSNA ICH}                                                                                                         & \multicolumn{2}{c}{ISIC 2019}               \\ \cmidrule(lr){3-16} 
				&                         & \multicolumn{2}{c}{$(\rho=0.2,   \eta=1.0)$} & \multicolumn{2}{c}{$(\rho=0.4, \eta=0.8)$} & \multicolumn{2}{c|}{$(\rho=0.6, \eta=0.5)$} & \multicolumn{2}{c}{$(\rho=0.1, \eta=1.0)$} & \multicolumn{2}{c}{$(\rho=0.2, \eta=0.8)$} & \multicolumn{2}{c|}{$(\rho=0.4, \eta=0.6)$} & \multicolumn{2}{c}{$(\rho=0.1, \eta=1.0)$} \\ \cmidrule(lr){3-16} 
				&                         & ACC                   & AUC                  & ACC                  & AUC                 & ACC                   & AUC                 & ACC                  & AUC                 & ACC                  & AUC                 & ACC                  & AUC                  & ACC                  & AUC                 \\ \midrule[1.2pt]
				Vanilla                                                                        & FedAvg                  & $10.94_{\pm 1.32}$    & $1.98_{\pm 0.41}$    & $12.36_{\pm 1.35}$   & $2.00_{\pm 0.29}$   & $14.56_{\pm 2.13}$    & $2.36_{\pm 0.29}$   & $5.07_{\pm 0.47}$    & $3.95_{\pm 0.27}$   & $5.09_{\pm 0.43}$    & $4.41_{\pm 0.41}$   & $4.26_{\pm 0.47}$    & $4.14_{\pm 0.41}$    & $15.01_{\pm 1.30}$   & $7.24_{\pm 1.14}$   \\ \midrule[1.2pt]
				\multirow{4}{*}{Fair FL}                                                       & q-FedAvg                & $10.63_{\pm 0.87}$    & $1.98_{\pm 0.33}$    & $10.06_{\pm 0.95}$   & $1.79_{\pm 0.22}$   & $7.42_{\pm 0.59}$     & $1.83_{\pm 0.40}$   & $4.49_{\pm 0.38}$    & $3.85_{\pm 0.31}$   & $4.44_{\pm 0.44}$    & $4.06_{\pm 0.41}$   & $3.42_{\pm 0.55}$    & $3.26_{\pm 0.40}$    & $11.36_{\pm 1.37}$   & $5.40_{\pm 0.70}$   \\
				& FedCE                   & $7.54_{\pm 1.89}$     & $2.03_{\pm 0.49}$    & $10.59_{\pm 1.75}$   & $2.29_{\pm 0.66}$   & $12.95_{\pm 1.41}$    & $2.69_{\pm 0.30}$   & $3.51_{\pm 0.31}$    & $3.36_{\pm 0.16}$   & $2.66_{\pm 1.37}$    & $3.85_{\pm 0.50}$   & $3.28_{\pm 0.45}$    & $3.38_{\pm 0.44}$    & $11.52_{\pm 0.76}$   & $5.05_{\pm 0.39}$   \\
				& FedGA                   & $14.84_{\pm 2.21}$    & $2.97_{\pm 1.32}$    & $11.73_{\pm 2.24}$   & $4.27_{\pm 0.59}$   & $11.29_{\pm 1.01}$    & $2.71_{\pm 0.34}$   & $2.62_{\pm 0.27}$    & $2.49_{\pm 0.26}$   & $\bm{2.18_{\pm 0.35}}$    & $2.28_{\pm 0.32}$   & $2.75_{\pm 0.37}$    & $2.54_{\pm 0.26}$    & $5.35_{\pm 0.75}$    & $2.58_{\pm 0.53}$   \\
				& FedISM                  & $9.39_{\pm 2.51}$     & $0.99_{\pm 0.19}$    & $5.41_{\pm 0.59}$    & $0.62_{\pm 0.08}$   & $3.74_{\pm 0.49}$     & $0.50_{\pm 0.06}$   & $2.77_{\pm 1.21}$    & $3.19_{\pm 0.20}$   & $4.28_{\pm 0.66}$    & $3.32_{\pm 0.22}$   & $2.85_{\pm 0.57}$    & $2.80_{\pm 0.27}$    & $13.02_{\pm 2.82}$   & $5.29_{\pm 0.74}$   \\ \midrule[1.2pt]
				\multirow{5}{*}{\begin{tabular}[c]{@{}l@{}}Fair FL +\\ Robust FL\end{tabular}} & q-FedAvg + Median       & $11.44_{\pm 1.16}$    & $1.81_{\pm 0.23}$    & $10.17_{\pm 1.95}$   & $1.78_{\pm 0.28}$   & $9.82_{\pm 1.09}$     & $2.05_{\pm 0.28}$   & $5.71_{\pm 0.75}$    & $4.49_{\pm 0.45}$   & $5.86_{\pm 0.91}$    & $5.49_{\pm 0.44}$   & $4.28_{\pm 0.45}$    & $4.26_{\pm 0.49}$    & $15.41_{\pm 2.03}$   & $7.35_{\pm 1.26}$   \\
				& FedCE + RFA             & $11.02_{\pm 1.56}$    & $2.02_{\pm 0.51}$    & $13.00_{\pm 2.29}$   & $1.95_{\pm 0.40}$   & $12.36_{\pm 1.30}$    & $2.16_{\pm 0.18}$   & $4.78_{\pm 0.51}$    & $4.14_{\pm 0.21}$   & $5.00_{\pm 0.43}$    & $4.39_{\pm 0.22}$   & $4.18_{\pm 0.46}$    & $4.09_{\pm 0.47}$    & $14.93_{\pm 1.22}$   & $7.00_{\pm 0.73}$   \\
				& FedGA + FedCorr         & $19.98_{\pm 7.19}$    & $2.86_{\pm 0.61}$    & $18.86_{\pm 3.62}$   & $4.11_{\pm 1.05}$   & $16.01_{\pm 6.37}$    & $2.80_{\pm 1.20}$   & $\bm{2.47_{\pm 0.97}}$    & $2.71_{\pm 0.21}$   & $3.92_{\pm 1.34}$    & $3.01_{\pm 0.40}$   & $4.96_{\pm 4.10}$    & $3.51_{\pm 1.65}$    & $22.40_{\pm 0.80}$   & $12.48_{\pm 0.75}$  \\
				& FedISM + FedNoRo        & $6.83_{\pm 0.44}$     & $0.80_{\pm 0.05}$    & $6.28_{\pm 0.42}$    & $0.75_{\pm 0.05}$   & $4.57_{\pm 0.39}$     & $0.61_{\pm 0.05}$   & $5.57_{\pm 2.55}$    & $3.73_{\pm 0.72}$   & $4.98_{\pm 2.30}$    & $3.41_{\pm 0.71}$   & $3.56_{\pm 0.17}$    & $3.54_{\pm 0.15}$    & $14.72_{\pm 0.57}$   & $6.86_{\pm 0.33}$   \\
				& H-nobs (Oracle)         & $8.72_{\pm 0.67}$     & $1.20_{\pm 0.11}$    & $6.50_{\pm 0.60}$    & $0.91_{\pm 0.07}$   & $3.85_{\pm 3.10}$     & $0.67_{\pm 0.47}$   & $3.76_{\pm 0.34}$    & $3.13_{\pm 0.21}$   & $3.40_{\pm 0.42}$    & $2.82_{\pm 0.18}$   & $2.92_{\pm 1.83}$    & $2.50_{\pm 0.94}$    & $8.60_{\pm 0.82}$    & $4.02_{\pm 0.46}$   \\ \midrule[1.2pt]
				\multirow{2}{*}{Ours}                                                          & FedPCA (D)              & $5.17_{\pm 0.61}$     & $0.62_{\pm 0.07}$    & $3.79_{\pm 0.71}$    & $0.48_{\pm 0.07}$   & $2.70_{\pm 0.34}$     & $0.39_{\pm 0.04}$   & $2.81_{\pm 0.18}$    & $2.45_{\pm 0.07}$   & $2.68_{\pm 0.20}$    & $2.33_{\pm 0.09}$   & $2.51_{\pm 0.17}$    & $2.17_{\pm 0.13}$    & $\bm{4.72_{\pm 1.09}}$    & $\bm{1.87_{\pm 0.51}}$   \\
				& FedPCA (HS)             & $\bm{4.88_{\pm 0.81}}$     & $\bm{0.60_{\pm 0.08}}$    & $\bm{3.08_{\pm 0.50}}$    & $\bm{0.44_{\pm 0.06}}$   & $\bm{1.59_{\pm 0.62}}$     & $\bm{0.25_{\pm 0.10}}$   & $2.60_{\pm 0.17}$    & $\bm{2.37_{\pm 0.07}}$   & $2.52_{\pm 0.23}$    & $\bm{2.17_{\pm 0.08}}$   & $\bm{2.17_{\pm 0.42}}$    & $\bm{1.79_{\pm 0.11}}$    & $4.85_{\pm 0.93}$    & $1.92_{\pm 0.47}$   \\ \bottomrule[2pt]
			\end{tabular}
		}
	\end{table*}

	\subsection{Analysis of Clients Prioritized by Different Methods}  \label{sec:appendix_clients_prioritized}
	
\begin{figure}[!h] 
		\centering
		\includegraphics[width=1.0\textwidth]{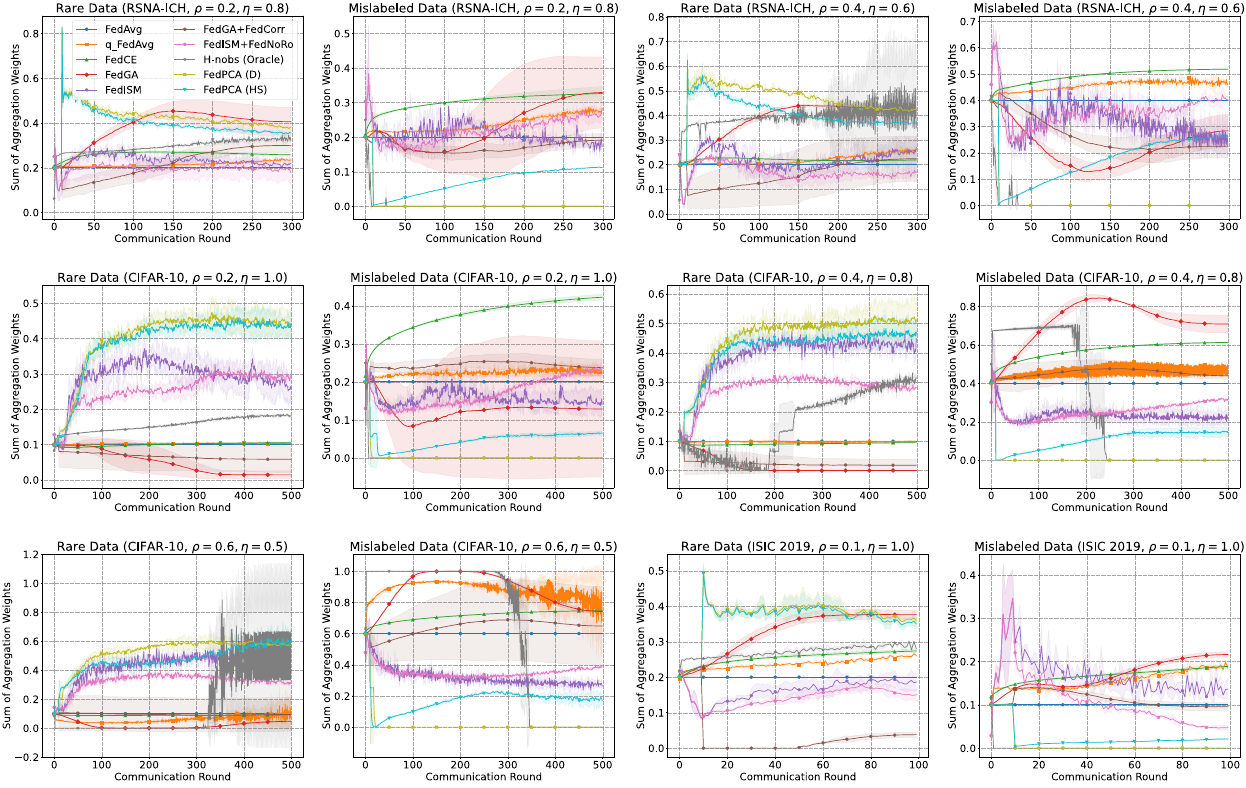}
		\caption{Aggregation weights assigned to clients over communication rounds, with the transparent area indicating the standard deviation. The legend in the first figure remains consistent across all figures.}
		\label{fig:appendix_more_weights}
	\end{figure}

	In the main text (see Fig. \ref{fig:weights}), we visualize client weights to analyze which clients are prioritized by different methods. Here, we provide additional results on other datasets and settings (see Fig. \ref{fig:appendix_more_weights}). These results demonstrate that our method consistently assigns higher weights to rare data while minimizing the impact of mislabeled data. In contrast, other methods fail to achieve this consistently, particularly in settings with a high proportion of mislabeled clients (\textit{e.g.}, CIFAR-10).
	
	To further quantify these observations, we conduct a quantitative analysis of the weights assigned to different client types across various methods. Specifically, we define the weights assigned to rare and mislabeled data clients as  $w_r$  and  $w_m$ , respectively. We then compare  $w_r(1 - w_m)$  across different methods, where higher values indicate a stronger emphasis on rare data while effectively suppressing mislabeled data, demonstrating an improved ability to jointly enhance fairness and robustness. The results, summarized in Tab. \ref{tab:appendix_weights}, show that our approach achieves the best performance, highlighting its superior capability in automatically identifying and down-weighting mislabeled data while amplifying the contribution of rare data clients. 
	
	\begin{table*}[!h]
		\centering
		\renewcommand{\arraystretch}{1.1}
		\caption{Comparison on $w_r(1 - w_m)$ averaged across communication rounds. Bold values highlight the best result.}
		\label{tab:appendix_weights}
		\resizebox{\textwidth}{!}{
			\begin{tabular}{ll|ccc|ccc|c}
				\toprule[2pt]
				\multirow{2}{*}{Category}                                                      & \multirow{2}{*}{Method} & \multicolumn{3}{c|}{CIFAR-10}                                              & \multicolumn{3}{c|}{RSNA ICH}                                            & ISIC 2019               \\ \cmidrule(lr){3-9} 
				&                         & $(\rho=0.2,   \eta=1.0)$ & $(\rho=0.4, \eta=0.8)$ & $(\rho=0.6, \eta=0.5)$ & $(\rho=0.1, \eta=1.0)$ & $(\rho=0.2, \eta=0.8)$ & $(\rho=0.4, \eta=0.6)$ & $(\rho=0.1, \eta=1.0)$ \\ \midrule[1.2pt]
				Vanilla                                                                        & FedAvg                  & $0.0800$                 & $0.0600$               & $0.0400$               & $0.1800$               & $0.1600$               & $0.1200$               & $0.1800$               \\ \midrule[1.2pt]
				\multirow{4}{*}{Fair FL}                                                       & q-FedAvg                & $0.0793$                 & $0.0531$               & $0.0106$               & $0.1922$               & $0.1636$               & $0.1194$               & $0.2005$               \\
				& FedCE                   & $0.0622$                 & $0.0389$               & $0.0253$               & $0.2182$               & $0.1816$               & $0.1113$               & $0.2122$               \\
				& FedGA                   & $0.0455$                 & $0.0068$               & $0.0053$               & $0.2562$               & $0.3001$               & $0.3025$               & $0.2715$               \\
				& FedISM                  & $0.2464$                 & $0.2997$               & $0.3041$               & $0.1595$               & $0.1842$               & $0.1506$               & $0.1369$               \\ \midrule[1.2pt]
				\multirow{3}{*}{\begin{tabular}[c]{@{}l@{}}Fair FL +\\ Robust FL\end{tabular}} & FedGA + FedCorr         & $0.0504$                 & $0.0140$               & $0.0600$               & $0.2048$               & $0.1817$               & $0.1238$               & $0.0310$               \\
				& FedISM + FedNoRo        & $0.2114$                 & $0.2090$               & $0.2122$               & $0.1558$               & $0.1523$               & $0.1113$               & $0.1287$               \\
				& H-nobs (Oracle)         & $0.1561$                 & $0.1494$               & $0.1584$               & $0.2692$               & $0.2953$               & $0.3907$               & $0.2713$               \\ \midrule[1.2pt]
				\multirow{2}{*}{Ours}                                                          & FedPCA (D)              & $\bm{0.4011}$            & $\bm{0.4510}$          & $\bm{0.5258}$          & $\bm{0.3954}$          & $\bm{0.4233}$          & $\bm{0.4669}$          & $\bm{0.3668}$          \\
				& FedPCA (HS)             & $0.3708$                 & $0.3685$               & $0.3900$               & $0.3733$               & $0.3699$               & $0.3431$               & $0.3558$               \\ \bottomrule[2pt]
			\end{tabular}
		}
	\end{table*}
	
	Please note that we exclude Median \cite{Median} and RFA \cite{RFA} due to their median-based aggregation, which complicates individual client weight quantification.

	\subsection{Discussion of Hyper-parameters}  \label{sec:appendix_experiments_discussion}
\begin{figure*}[!h] 
		\centering
		\includegraphics[width=1.0\textwidth]{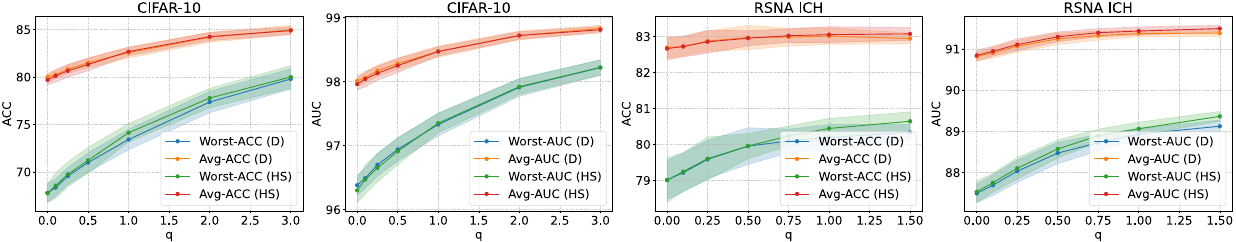}
		\caption{The impact of $q$ on ACC and AUC for our solutions, \textit{i.e.}, FedPCA (D) and FedPCA (HS), on two datasets. We set $\rho$ and $\eta$ as (0.1, 1.0) and (0.2, 1.0) for RSNA ICH and CIFAR-10, respectively.}
		\label{fig:appendix_q}
	\end{figure*}

	\noindent\textbf{Impact of $q$.}
	$q$ is a key parameter in Eq. \ref{eq:weights}, controlling the prioritization of clients with lower dispersion scores. To analyze its impact, we vary $q$ from the default setting for CIFAR-10 and RSNA ICH, and show the resulting performance changes in Fig. \ref{fig:appendix_q}. The results indicate that larger $q$ values benefit the learning of the worst-performing distribution, as reflected by increases in both ACC and AUC. Additionally, while the average ACC and AUC also improve, the gains are slower compared to those for the worst distribution, suggesting a trade-off where the best-performing distribution may be slightly compromised.

\begin{figure*}[!h] 
		\centering
		\includegraphics[width=1.0\textwidth]{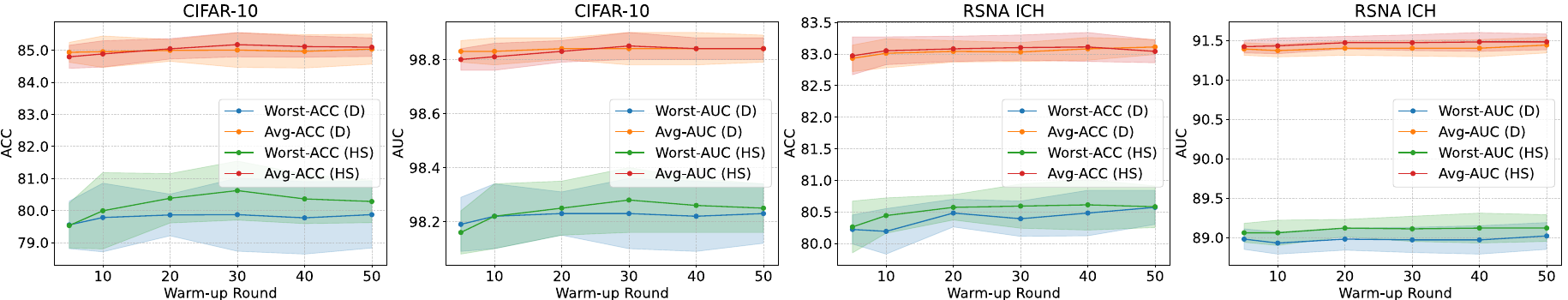}
		\caption{The impact of warm-up rounds on ACC and AUC for our solutions, \textit{i.e.}, FedPCA (D) and FedPCA (HS), on two datasets. We set $\rho$ and $\eta$ as (0.1, 1.0) and (0.2, 1.0) for RSNA ICH and CIFAR-10, respectively.}
		\label{fig:appendix_wr}
	\end{figure*}

	\vspace{0.1cm}
	\noindent\textbf{Impact of Warm-up Rounds.}
	We initialize the model with FedAvg \cite{FedAvg} during a warm-up phase before applying our solution, following standard practices for learning with noisy labels \cite{FedNoRo}. As shown in Fig. \ref{fig:appendix_wr}, varying the number of warm-up rounds has minimal influence on the final performance, reducing the need for extensive tuning of this parameter in our method.

	\subsection{Visualization of Feature Space}  \label{sec:appendix_experiments_visualization}
\begin{figure*}[!t] 
		\centering
		\includegraphics[width=0.9\textwidth]{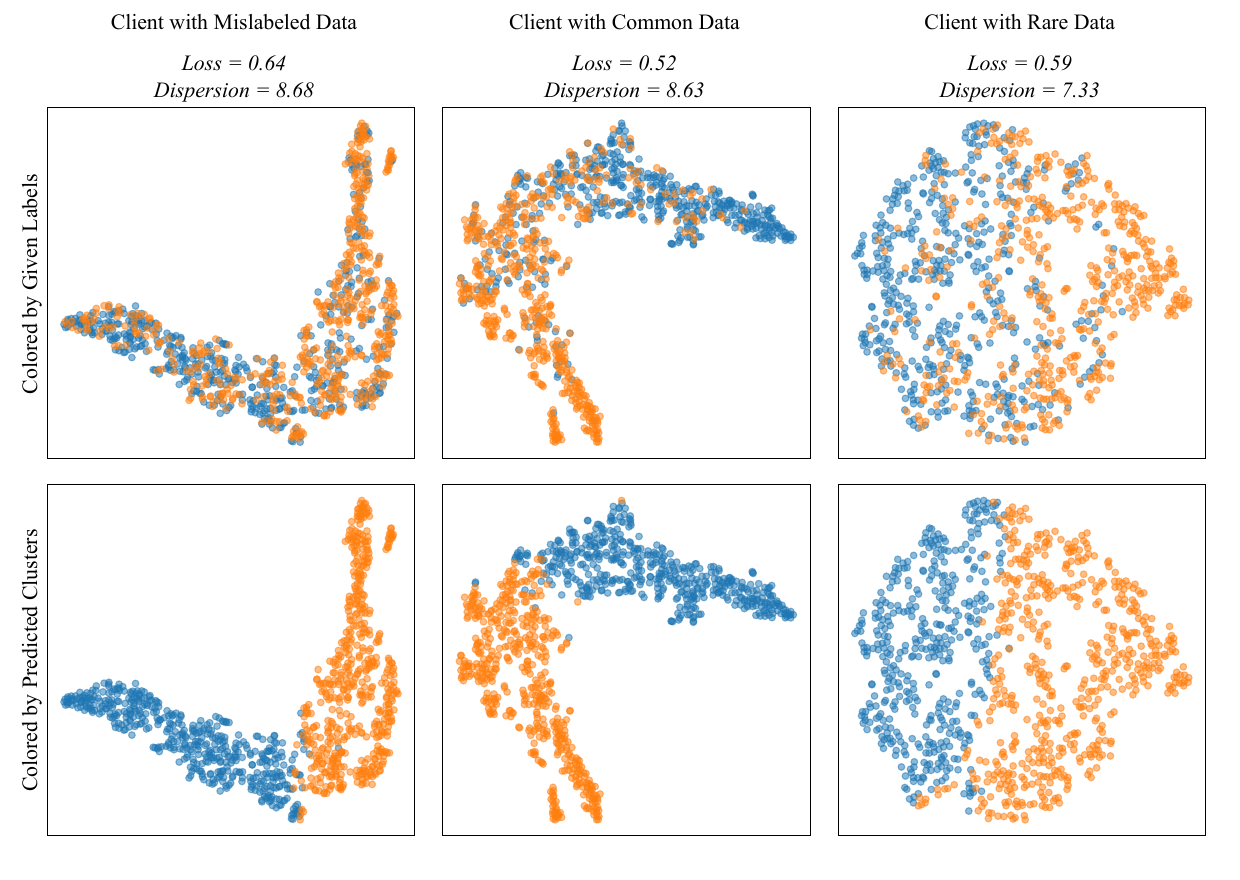}
		\caption{TSNE visualization of feature space in different clients.}
		\label{fig:tsne}
	\end{figure*}

	We visualize the feature distributions of different clients using the widely adopted t-SNE technique, as shown in Fig. \ref{fig:tsne}. Examining loss reveals that both rare and mislabeled data clients exhibit higher loss compared to common data clients due to a significant discrepancy between labels and predictions. However, their feature spaces differ fundamentally: mislabeled data clients form distinct clusters, while rare data clients lack such separation, with centroids positioned closely together (see the second row). This distinction is effectively captured by the dispersion score. By integrating these two metrics, our proposed performance-capacity analysis accurately identifies mislabeled clients based on their unexpectedly high loss relative to their dispersion score.

	\subsection{Results under two More General Client Data Distribution Settings} \label{sec:appendix_experiments_data_dirtibution}
	In the main text, we generate common and rare data clients by first randomly partitioning the dataset into multiple subsets, followed by corrupting a minority of clients. This approach assumes that all data within each client is \textit{i.i.d.} sampled from either the common data distribution $\mathcal{P}_1$ or the rare data distribution $\mathcal{P}_2$. Here, we explore two more general scenarios created by non-\textit{i.i.d.} sampling strategies to evaluate whether the methods can robustly achieve fairness.

	\vspace{0.1cm}
	\noindent\textbf{Non-\textit{i.i.d.} Label Distribution.}
	Non-\textit{i.i.d.} label distribution, where $P_i(y) \not\equiv P_j(y)$ for $i \neq j$, is common in real-world applications and widely studied in FL. To simulate this setting, we first partition the dataset using a Dirichlet distribution (\textit{i.e.}, Dir(2.0)) to introduce heterogeneous class distributions across clients, after which we apply corruption to a subset of minority clients. This setup results in clients with imbalanced and heterogeneous class distributions, further challenging client identification and the application of fair and robust strategies. We conduct experiments on CIFAR-10, where all methods are combined with logit adjustment \cite{menonlong} to mitigate label skew. Additionally, our dispersion score is normalized by each client's data size to ensure comparability. All other settings remain consistent with the main text.
	The results, summarized in Tab. \ref{tab:appendix_SOTA_CIFAR_label_noniid}, demonstrate that our method consistently outperforms others. This underscores the broad applicability of our solution.

	\begin{table*}[!h]
		\centering
		\renewcommand{\arraystretch}{1.1}
		\caption{Performance comparison on CIFAR-10 (mean (\%) $\pm$ standard deviation (\%)) under a non-\textit{i.i.d.} label distribution. Bold values indicate the best result.}
		\label{tab:appendix_SOTA_CIFAR_label_noniid}
		\resizebox{\textwidth}{!}{
			\begin{tabular}{ll|cccc|cccc|cccc}
				\toprule[2pt]
				\multirow{4}{*}{Category}                                                      & \multirow{4}{*}{Method} & \multicolumn{4}{c|}{$(\rho=0.1, \eta=1.0)$}                                        & \multicolumn{4}{c|}{$(\rho=0.2, \eta=0.8)$}                                       & \multicolumn{4}{c}{$(\rho=0.4, \eta=0.6)$}                                        \\ \cmidrule(lr){3-14}
				&                         & \multicolumn{2}{c}{Worst}                & \multicolumn{2}{c|}{Average}            & \multicolumn{2}{c}{Worst}               & \multicolumn{2}{c|}{Average}            & \multicolumn{2}{c}{Worst}               & \multicolumn{2}{c}{Average}             \\ \cmidrule(lr){3-14}
				&                         & ACC                 & AUC                & ACC                & AUC                & ACC                & AUC                & ACC                & AUC                & ACC                & AUC                & ACC                & AUC                \\ \midrule[1.2pt]
				Vanilla                                                                        & FedAvg                  & $63.34_{\pm 2.84}$  & $94.82_{\pm 0.92}$ & $75.15_{\pm 2.36}$ & $96.82_{\pm 0.68}$ & $64.05_{\pm 5.12}$ & $95.43_{\pm 0.91}$ & $75.01_{\pm 3.13}$ & $97.08_{\pm 0.59}$ & $56.90_{\pm 2.86}$ & $94.80_{\pm 0.43}$ & $70.50_{\pm 1.54}$ & $96.73_{\pm 0.24}$ \\ \midrule[1.2pt]
				\multirow{4}{*}{Fair FL}                                                       & q-FedAvg                & $65.09_{\pm 4.22}$  & $95.01_{\pm 1.05}$ & $76.38_{\pm 2.81}$ & $96.96_{\pm 0.70}$ & $61.87_{\pm 4.96}$	  & $94.90_{\pm 0.97}$	  & $73.26_{\pm 2.99}$	  & $96.69_{\pm 0.65}$	  & $63.07_{\pm 3.15}$	  & $93.73_{\pm 1.15}$	  & $70.76_{\pm 2.76}$	  & $95.46_{\pm 0.92}$\\
				& FedCE                   & $60.25_{\pm 5.63}$  & $92.24_{\pm 2.94}$ & $67.94_{\pm 7.19}$ & $94.26_{\pm 2.51}$ & $48.87_{\pm 7.73}$ & $92.57_{\pm 1.42}$ & $61.32_{\pm 7.03}$ & $94.69_{\pm 1.22}$ & $59.41_{\pm 2.98}$ & $94.43_{\pm 0.49}$ & $70.35_{\pm 1.75}$ & $96.35_{\pm 0.29}$ \\
				& FedGA                   & $58.48_{\pm 2.56}$  & $94.33_{\pm 0.69}$ & $71.37_{\pm 2.03}$ & $96.48_{\pm 0.55}$ & $53.87_{\pm 6.60}$ & $91.69_{\pm 1.80}$ & $63.84_{\pm 5.29}$ & $94.01_{\pm 1.41}$ & $56.53_{\pm 2.17}$ & $93.14_{\pm 0.39}$ & $68.26_{\pm 1.40}$ & $95.44_{\pm 0.19}$ \\
				& FedISM                  & $63.13_{\pm 4.28}$  & $96.53_{\pm 0.37}$ & $75.17_{\pm 2.79}$ & $97.86_{\pm 0.21}$ & $73.24_{\pm 1.65}$ & $97.47_{\pm 0.16}$ & $80.10_{\pm 1.10}$ & $98.25_{\pm 0.11}$ & $76.73_{\pm 1.43}$ & $97.64_{\pm 0.13}$ & $81.35_{\pm 0.72}$ & $98.26_{\pm 0.08}$ \\ \midrule[1.2pt]
				\multirow{5}{*}{\begin{tabular}[c]{@{}l@{}}Fair FL +\\ Robust FL\end{tabular}} & q-FedAvg + Median       & $63.94_{\pm 3.13}$  & $95.48_{\pm 0.33}$ & $75.79_{\pm 2.29}$ & $97.32_{\pm 0.23}$ & $64.38_{\pm 3.97}$ & $95.49_{\pm 0.69}$ & $75.48_{\pm 2.56}$ & $97.16_{\pm 0.45}$ & $66.98_{\pm 1.68}$ & $95.59_{\pm 0.24}$ & $77.06_{\pm 0.88}$ & $97.23_{\pm 0.16}$ \\
				& FedCE + RFA             & $64.33_{\pm 3.18}$  & $95.27_{\pm 1.05}$ & $75.15_{\pm 3.34}$ & $97.02_{\pm 0.81}$ & $56.79_{\pm 6.62}$ & $95.04_{\pm 0.88}$ & $69.58_{\pm 4.52}$ & $96.73_{\pm 0.62}$ & $62.09_{\pm 4.02}$ & $95.11_{\pm 0.69}$ & $73.79_{\pm 2.12}$ & $96.96_{\pm 0.36}$ \\
				& FedGA + FedCorr         & $48.24_{\pm 11.23}$ & $93.27_{\pm 2.41}$ & $66.83_{\pm 6.89}$ & $95.85_{\pm 1.49}$ & $42.22_{\pm 8.52}$ & $91.66_{\pm 1.59}$ & $60.88_{\pm 4.87}$ & $94.57_{\pm 0.94}$ & $54.79_{\pm 5.55}$ & $93.62_{\pm 0.63}$ & $67.99_{\pm 2.63}$ & $95.72_{\pm 0.33}$ \\
				& FedISM + FedNoRo        & $74.07_{\pm 1.24}$  & $97.47_{\pm 0.10}$ & $82.15_{\pm 0.67}$ & $98.44_{\pm 0.05}$ & $74.99_{\pm 0.91}$ & $97.51_{\pm 0.12}$ & $82.00_{\pm 0.49}$ & $98.40_{\pm 0.06}$ & $77.10_{\pm 0.76}$ & $97.70_{\pm 0.08}$ & $82.31_{\pm 0.40}$ & $98.41_{\pm 0.04}$ \\
				& H-nobs (Oracle)         & $74.56_{\pm 1.37}$  & $97.18_{\pm 0.23}$ & $83.34_{\pm 0.64}$ & $98.38_{\pm 0.11}$ & $74.88_{\pm 1.84}$	  & $96.89_{\pm 0.38}$	  & $81.59_{\pm 0.90}$	  & $97.99_{\pm 0.21}$	  & $60.50_{\pm 2.51}$	  & $92.01_{\pm 0.76}$	  & $66.05_{\pm 1.65}$	  & $93.52_{\pm 0.74}$\\ \midrule[1.2pt]
				\multirow{2}{*}{Ours}                                                          & FedPCA (D)              & $79.66_{\pm 1.05}$      & $98.15_{\pm 0.10}$      & $84.44_{\pm 0.32}$      & $\bm{98.74_{\pm 0.03}}$ & $\bm{80.73_{\pm 0.97}}$ & $\bm{98.19_{\pm 0.10}}$ & $\bm{84.25_{\pm 0.32}}$ & $\bm{98.68_{\pm 0.04}}$ & $\bm{79.87_{\pm 0.67}}$ & $\bm{98.01_{\pm 0.08}}$ & $\bm{83.21_{\pm 0.27}}$ & $\bm{98.51_{\pm 0.03}}$ \\
				& FedPCA (HS)             & $\bm{80.02_{\pm 0.84}}$ & $\bm{98.16_{\pm 0.08}}$ & $\bm{84.54_{\pm 0.35}}$ & $\bm{98.74_{\pm 0.04}}$ & $80.06_{\pm 0.99}$      & $98.14_{\pm 0.09}$      & $82.96_{\pm 0.86}$      & $98.58_{\pm 0.05}$      & $79.25_{\pm 0.86}$      & $97.93_{\pm 0.11}$      & $81.28_{\pm 0.58}$      & $98.32_{\pm 0.04}$      \\ \bottomrule[2pt]
			\end{tabular}
		}
	\end{table*}

	\vspace{0.1cm}
	\noindent\textbf{Non-\textit{i.i.d.} Input Distribution.}
	We also consider a more generalized scenario where input images are not strictly \textit{i.i.d.} from the common data distribution $\mathcal{P}_1$ or the rare data distribution $\mathcal{P}_2$, but instead follow a mixed distribution $\alpha \mathcal{P}_1 + (1-\alpha)\mathcal{P}_2$, where $\alpha \in [0,1]$ represents the proportion of common data. To evaluate method effectiveness under this setting, we extend the CIFAR-10 experiments by modifying the five rare clients from Sec. \ref{sec:setup} to contain mixed data. Their common data proportions $\alpha$ are set to 0.8, 0.6, 0.4, 0.2, and 0.0, with the remaining data being rare. This distribution introduces challenges for clustering, as mixed distributions may be ambiguously assigned to $\mathcal{S}_c$ or $\mathcal{S}_r$.
	
\begin{figure}[!h] 
		\centering
		\includegraphics[width=0.35\textwidth]{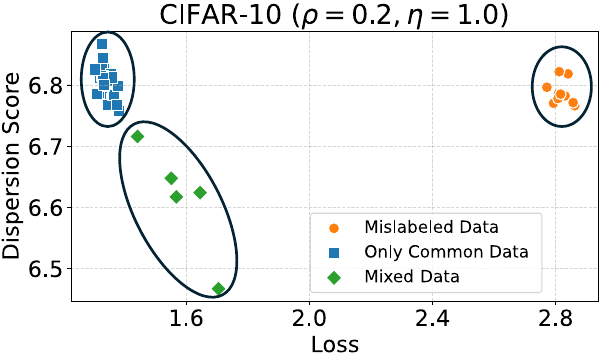}
		\caption{Visualization of performance-capacity analysis under the non-\textit{i.i.d.} input setting. Ellipses highlight clusters identified by Gaussian mixture models.}
		\label{fig:appendix_mixed}
	\end{figure}

	As shown in Fig. \ref{fig:appendix_mixed}, correctly labeled clients exhibit a clear inverse relationship between the two metrics, whereas mislabeled clients deviate and cluster in the upper right. This is consistent as what we discuss in the main text. While rare-data clients with low common data proportions may occasionally be grouped with common-data clients, this does not impact performance, as learning after mislabeled client identification is driven by the dispersion score rather than clustering (see Eq. \ref{eq:weights}). The comparison results under this setting, summarized in Tab. \ref{tab:appendix_SOTA_CIFAR_input_noniid}, demonstrate that our solution consistently outperforms other methods.

	\begin{table*}[!h]
		\centering
		\renewcommand{\arraystretch}{1.1}
		\caption{Performance comparison on CIFAR-10 (mean (\%) $\pm$ standard deviation (\%)) under a non-\textit{i.i.d.} input distribution. Bold values indicate the best result.}
		\label{tab:appendix_SOTA_CIFAR_input_noniid}
		\resizebox{\textwidth}{!}{
			\begin{tabular}{ll|cccc|cccc|cccc}
				\toprule[2pt]
				\multirow{4}{*}{Category}                                                      & \multirow{4}{*}{Method} & \multicolumn{4}{c|}{$(\rho=0.1, \eta=1.0)$}                                        & \multicolumn{4}{c|}{$(\rho=0.2, \eta=0.8)$}                                         & \multicolumn{4}{c}{$(\rho=0.4, \eta=0.6)$}                                        \\ \cmidrule(lr){3-14}
				&                         & \multicolumn{2}{c}{Worst}                & \multicolumn{2}{c|}{Average}            & \multicolumn{2}{c}{Worst}                & \multicolumn{2}{c|}{Average}             & \multicolumn{2}{c}{Worst}               & \multicolumn{2}{c}{Average}             \\ \cmidrule(lr){3-14}
				&                         & ACC                 & AUC                & ACC                & AUC                & ACC                 & AUC                & ACC                 & AUC                & ACC                & AUC                & ACC                & AUC                \\ \midrule[1.2pt]
				Vanilla                                                                        & FedAvg                  & $72.58_{\pm 3.75}$  & $96.39_{\pm 0.88}$ & $80.22_{\pm 3.43}$ & $97.68_{\pm 0.66}$ & $67.53_{\pm 3.66}$  & $95.00_{\pm 1.34}$ & $75.98_{\pm 3.04}$  & $96.63_{\pm 0.99}$ & $66.99_{\pm 6.27}$ & $95.77_{\pm 0.70}$ & $76.34_{\pm 3.85}$ & $97.26_{\pm 0.41}$ \\ \midrule[1.2pt]
				\multirow{4}{*}{Fair FL}                                                       & q-FedAvg                & $71.57_{\pm 3.93}$  & $95.74_{\pm 1.19}$ & $78.98_{\pm 3.65}$ & $97.18_{\pm 0.91}$ & $69.80_{\pm 3.13}$	  & $95.38_{\pm 1.11}$	  & $77.20_{\pm 2.90}$	  & $96.85_{\pm 0.87}$	  & $64.85_{\pm 4.54}$	  & $93.10_{\pm 2.32}$	  & $71.20_{\pm 4.98}$	  & $94.79_{\pm 2.00}$ \\
				& FedCE                   & $64.53_{\pm 8.75}$  & $93.62_{\pm 2.74}$ & $71.00_{\pm 9.61}$ & $95.30_{\pm 2.34}$ & $48.11_{\pm 12.91}$ & $92.11_{\pm 2.77}$ & $57.69_{\pm 12.80}$ & $94.11_{\pm 2.35}$ & $62.80_{\pm 2.20}$ & $94.33_{\pm 0.46}$ & $72.00_{\pm 1.47}$ & $96.23_{\pm 0.28}$ \\
				& FedGA                   & $59.26_{\pm 13.86}$ & $93.23_{\pm 3.66}$ & $73.65_{\pm 9.00}$ & $96.10_{\pm 2.28}$ & $39.98_{\pm 7.12}$  & $84.54_{\pm 3.56}$ & $50.55_{\pm 8.52}$  & $88.69_{\pm 3.18}$ & $58.94_{\pm 2.82}$ & $92.63_{\pm 1.04}$ & $69.18_{\pm 1.79}$ & $95.13_{\pm 0.61}$ \\
				& FedISM                  & $73.73_{\pm 1.99}$  & $97.69_{\pm 0.23}$ & $80.98_{\pm 1.71}$ & $98.50_{\pm 0.17}$ & $77.56_{\pm 1.50}$  & $97.90_{\pm 0.10}$ & $82.79_{\pm 1.03}$  & $98.56_{\pm 0.07}$ & $80.60_{\pm 1.00}$ & $98.06_{\pm 0.12}$ & $84.32_{\pm 0.46}$ & $98.62_{\pm 0.06}$ \\ \midrule[1.2pt]
				\multirow{5}{*}{\begin{tabular}[c]{@{}l@{}}Fair FL +\\ Robust FL\end{tabular}} & q-FedAvg + Median       & $71.88_{\pm 5.70}$  & $96.54_{\pm 0.91}$ & $79.77_{\pm 4.95}$ & $97.82_{\pm 0.66}$ & $70.47_{\pm 3.24}$  & $95.93_{\pm 0.92}$ & $77.62_{\pm 3.00}$  & $97.22_{\pm 0.74}$ & $71.42_{\pm 1.80}$ & $96.27_{\pm 0.25}$ & $79.28_{\pm 1.09}$ & $97.57_{\pm 0.16}$ \\
				& FedCE + RFA             & $64.94_{\pm 7.57}$  & $95.05_{\pm 1.63}$ & $73.67_{\pm 7.44}$ & $96.70_{\pm 1.26}$ & $63.25_{\pm 9.08}$  & $95.10_{\pm 1.46}$ & $72.04_{\pm 7.66}$  & $96.58_{\pm 1.15}$ & $69.28_{\pm 2.35}$ & $95.88_{\pm 0.38}$ & $77.53_{\pm 1.43}$ & $97.32_{\pm 0.22}$ \\
				& FedGA + FedCorr         & $48.88_{\pm 9.21}$  & $92.15_{\pm 2.56}$ & $67.41_{\pm 4.98}$ & $95.28_{\pm 1.52}$ & $41.85_{\pm 12.34}$ & $90.34_{\pm 2.19}$ & $61.67_{\pm 6.89}$  & $94.02_{\pm 1.35}$ & $46.81_{\pm 6.04}$ & $90.45_{\pm 1.44}$ & $62.76_{\pm 3.60}$ & $93.82_{\pm 0.90}$ \\
				& FedISM + FedNoRo        & $79.08_{\pm 0.63}$  & $98.09_{\pm 0.09}$ & $84.91_{\pm 0.32}$ & $98.80_{\pm 0.04}$ & $79.64_{\pm 1.12}$  & $98.10_{\pm 0.11}$ & $84.88_{\pm 0.67}$  & $98.76_{\pm 0.06}$ & $80.96_{\pm 0.51}$ & $98.18_{\pm 0.05}$ & $85.15_{\pm 0.22}$ & $98.75_{\pm 0.03}$ \\
				& H-nobs (Oracle)         & $81.01_{\pm 0.54}$  & $98.08_{\pm 0.07}$ & $86.67_{\pm 0.27}$ & $98.84_{\pm 0.04}$ & $81.24_{\pm 0.37}$	  & $98.05_{\pm 0.07}$	  & $86.16_{\pm 0.21}$	  & $98.74_{\pm 0.04}$	  & $77.92_{\pm 2.69}$	  & $97.50_{\pm 0.43}$	  & $82.31_{\pm 1.63}$	  & $98.22_{\pm 0.24}$ \\ \midrule[1.2pt]
				\multirow{2}{*}{Ours}                                                          & FedPCA (D)              & $\bm{82.28_{\pm 0.64}}$ & $\bm{98.46_{\pm 0.08}}$ & $\bm{86.91_{\pm 0.35}}$ & $\bm{99.02_{\pm 0.04}}$ & $\bm{82.46_{\pm 0.46}}$ & $\bm{98.43_{\pm 0.06}}$ & $\bm{86.56_{\pm 0.24}}$ & $\bm{98.95_{\pm 0.03}}$ & $\bm{81.94_{\pm 0.54}}$ & $\bm{98.27_{\pm 0.03}}$ & $\bm{85.39_{\pm 0.34}}$ & $\bm{98.78_{\pm 0.02}}$ \\
				& FedPCA (HS)             & $82.07_{\pm 0.63}$  & $98.43_{\pm 0.07}$ & $86.66_{\pm 0.33}$ & $98.99_{\pm 0.03}$ & $82.09_{\pm 0.48}$  & $98.42_{\pm 0.04}$ & $85.77_{\pm 0.36}$  & $98.91_{\pm 0.02}$ & $80.09_{\pm 0.69}$ & $98.06_{\pm 0.09}$ & $83.34_{\pm 0.66}$ & $98.56_{\pm 0.04}$ \\ \bottomrule[2pt]
			\end{tabular}
		}
	\end{table*}

	\subsection{Results under a More General Label Noise Distribution Setting} \label{sec:appendix_experiments_noise_dirtibution}
	In the main text, mislabeled data clients are generated by flipping ground-truth labels from previously created common data clients, essentially assuming that label noise is concentrated in distributions with larger data volumes. This assumption is grounded in real-world observations. In practice, rare data is valued for its uniqueness and limited availability, often receiving meticulous attention to ensure accurate labeling. In contrast, common data, held by the majority of clients, deviates from this ideal. The sheer volume of common data makes high-quality labeling both costly and impractical \cite{han2018co}, and its ubiquity often results in reduced oversight and safeguards, making it more susceptible to label poisoning attacks. These factors collectively contribute to the presence of label noise among clients with common data.
	
\begin{figure}[!h] 
		\centering
		\includegraphics[width=0.37\textwidth]{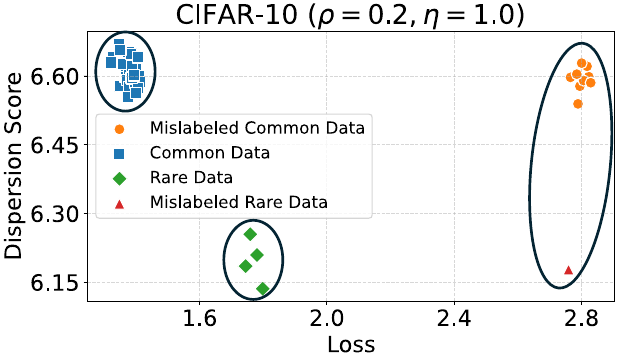}
		\caption{Visualization of performance-capacity analysis under the uniform label noise distribution setting. Ellipses highlight clusters identified by Gaussian mixture models.}
		\label{fig:appendix_noise}
	\end{figure}

	To further evaluate robustness, we also consider a more general noise distribution in which label noise is uniformly distributed across all clients. We conduct experiments on CIFAR-10 with $\rho = 0.2$ and $\eta = 1.0$, while keeping other settings consistent with the main text. This results in 9 common data clients and 1 rare data client being affected by label noise. Fig. \ref{fig:appendix_noise} illustrates the separability of the four client types under our proposed performance-capacity analysis. Notably, clients with mislabeled data consistently appear in the upper-right region of the plane, exhibiting unexpectedly high loss relative to their dispersion scores. This ensures precise identification, as demonstrated in Fig. \ref{fig:appendix_noise}, where clustering effectively distinguishes these clients. The comparison results in Tab. \ref{tab:appendix_SOTA_CIFAR_noise} further highlight the superiority of our approach in addressing this more generalized setting.
	
	\begin{table*}[!h]
		\centering
		\renewcommand{\arraystretch}{1.1}
		\caption{Performance comparison on CIFAR-10 (mean (\%) $\pm$ standard deviation (\%)) under the uniform label noise distribution. Bold values indicate the best result.}
		\label{tab:appendix_SOTA_CIFAR_noise}
		\resizebox{0.6\textwidth}{!}{
			\begin{tabular}{ll|cccc}
				\toprule[2pt]
				\multirow{3}{*}{Category}                                                      & \multirow{3}{*}{Method} & \multicolumn{4}{c}{$(\rho=0.1, \eta=1.0)$}                                         \\ \cmidrule(lr){3-6} 
				&                         & \multicolumn{2}{c}{Worst}                & \multicolumn{2}{c}{Average}             \\ \cmidrule(lr){3-6} 
				&                         & ACC                 & AUC                & ACC                & AUC                \\ \midrule[1.2pt]
				Vanilla                                                                        & FedAvg                  & $61.44_{\pm 6.58}$  & $93.96_{\pm 1.55}$ & $74.10_{\pm 4.98}$ & $96.40_{\pm 0.99}$ \\ \midrule[1.2pt]
				\multirow{4}{*}{Fair FL}                                                       & q-FedAvg                & $60.37_{\pm 5.46}$  & $93.15_{\pm 1.80}$ & $73.17_{\pm 4.37}$ & $95.86_{\pm 1.20}$ \\
				& FedCE                   & $59.59_{\pm 8.27}$  & $92.45_{\pm 2.80}$ & $69.26_{\pm 8.52}$ & $94.85_{\pm 2.23}$ \\
				& FedGA                   & $58.97_{\pm 9.48}$  & $92.92_{\pm 3.65}$ & $74.09_{\pm 7.33}$ & $95.98_{\pm 2.32}$ \\
				& FedISM                  & $63.50_{\pm 4.11}$  & $96.72_{\pm 0.33}$ & $75.64_{\pm 2.54}$ & $97.96_{\pm 0.20}$ \\ \midrule[1.2pt]
				\multirow{5}{*}{\begin{tabular}[c]{@{}l@{}}Fair FL +\\ Robust FL\end{tabular}} & q-FedAvg + Median       & $59.93_{\pm 7.83}$  & $94.48_{\pm 1.31}$ & $73.06_{\pm 6.62}$ & $96.75_{\pm 0.88}$ \\
				& FedCE + RFA             & $58.28_{\pm 7.63}$  & $93.59_{\pm 1.87}$ & $71.45_{\pm 6.69}$ & $96.08_{\pm 1.32}$ \\
				& FedGA + FedCorr         & $48.90_{\pm 10.82}$ & $89.89_{\pm 3.54}$ & $64.06_{\pm 8.16}$ & $93.41_{\pm 2.44}$ \\
				& FedISM + FedNoRo        & $74.08_{\pm 1.17}$  & $97.56_{\pm 0.14}$ & $82.15_{\pm 0.63}$ & $98.51_{\pm 0.07}$ \\
				& H-nobs (Oracle)         & $72.88_{\pm 0.78}$  & $96.80_{\pm 0.14}$ & $82.54_{\pm 0.41}$ & $98.20_{\pm 0.07}$ \\ \midrule[1.2pt]
				\multirow{2}{*}{Ours}                                                          & FedPCA (D)              & $\bm{78.28_{\pm 0.68}}$  & $\bm{98.02_{\pm 0.10}}$ & $\bm{84.33_{\pm 0.36}}$ & $\bm{98.74_{\pm 0.04}}$ \\
				& FedPCA (HS)             & $76.71_{\pm 0.93}$  & $97.82_{\pm 0.11}$ & $82.23_{\pm 0.84}$ & $98.57_{\pm 0.07}$ \\ \bottomrule[2pt]
			\end{tabular}
		}
	\end{table*}

\end{document}